\documentclass[10pt,letterpaper,twocolumn]{article}

\usepackage[T1]{fontenc}
\usepackage{lmodern}
\usepackage[margin=0.75in]{geometry}
\usepackage[hyphens]{url}
\usepackage{graphicx}
\usepackage{natbib}
\usepackage{caption}
\usepackage{algorithm}
\usepackage{algorithmic}
\usepackage{amsmath}
\usepackage{amssymb}
\usepackage{booktabs}
\usepackage{float}
\usepackage{placeins}
\usepackage{microtype}
\usepackage[colorlinks=true,linkcolor=blue,citecolor=blue,urlcolor=blue]{hyperref}

\title{Self-Verifier Capture in Unified Visual Generators:\\
Diagnosing and Repairing Closed-Loop Self-Reward}

\author{Binyan Chen \quad Tong Pan\\
University of Science and Technology of China\\
\texttt{sztn@mail.ustc.edu.cn} \quad \texttt{pantong@mail.ustc.edu.cn}}
\date{}

\hypersetup{
  pdftitle={Self-Verifier Capture in Unified Visual Generators: Diagnosing and Repairing Closed-Loop Self-Reward},
  pdfauthor={Binyan Chen and Tong Pan}
}

\begin{document}

\maketitle

\begin{abstract}
Image-space reward requires decoding and a separate judge; unified
visual generators can instead reuse their understanding pathway for
decode-free, self-supplied reward.  Optimizing against this verifier
closes the self-reward loop; shared trainable weights let one update
move generation and scoring even with the reward gradient stopped.  We
call the resulting failure \emph{self-verifier capture}: self-reward
rises while independently measured quality falls.  Matched
semantic/denoising $\times$ shared/frozen cells on Scale-RAE isolate two
routes: the policy exploits on-manifold blind spots of a frozen semantic
judge, while shared updates additionally rewrite the judge.  A frozen
judge also becomes stale, and repeated self-refresh yields progressively
weaker recoveries.  The diagnosis yields a two-part repair: locking the
warmed reward read blocks rewriting, while sparse, parameter-disjoint
external anchors counter residual exploitation.  Unanchored rollouts
retain semantic reward across multiple latent noise levels---dense
latent reward that terminal image-space judging
cannot supply---while a drift-triggered controller allocates anchors to
high-drift rollouts and propagates their calibration within each group.
At matched budgets, the controller improves on fixed-dose anchoring,
increasingly so as coverage tightens.  Both the unanchored failure and
the repair transfer to the higher-performing, architecturally distinct
Show-o2.  Across both bases, the complete method significantly
outperforms a construct-matched every-rollout judge on GenEval and
DPG-Bench while externally scoring only $1/16$ of rollouts.  At $1/32$
coverage, it scores higher in all four comparisons and remains jointly
non-inferior.
\end{abstract}

\section{Introduction}
\label{sec:introduction}

Unified visual generators increasingly place understanding and
generation in a shared model and latent space
\citep{ShowO,JanusFlow,ShowO2,ScaleRAE}.  Their understanding pathway can supply
decode-free self-reward without rendering pixels or training a
separate judge.  Yet the same trainable sharing couples policy and
judge: optimization can change both the generated samples and the
function that scores them.

\begin{figure}[!t]
  \centering
  \includegraphics[width=\columnwidth]{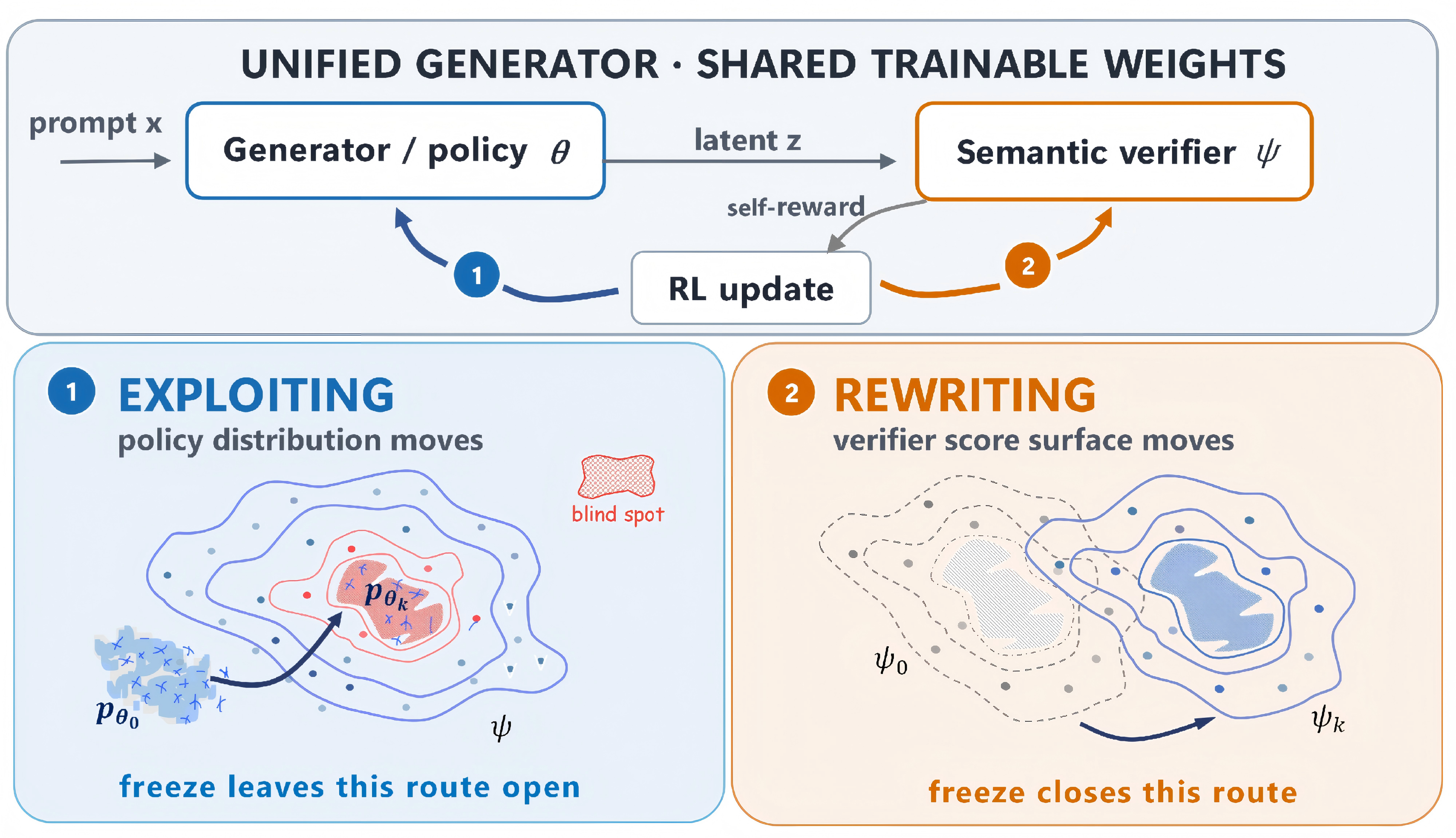}
  \caption{Two routes through closed-loop self-reward.  A policy
  update can search a fixed verifier's blind spots (blue,
  \emph{exploiting}); with trainable weight sharing, the same update
  can also move the verifier (orange, \emph{rewriting}).}
  \label{fig:closed-loop}
\end{figure}

Learned proxies are known to become unreliable when optimized beyond
their fitting regime \citep{GaoRewardOveroptimization}, but the usual
proxy-overoptimization picture treats the judge as fixed.  A unified
generator instead closes the reward loop.  We call its characteristic
failure \emph{self-verifier capture}: optimization raises self-reward
while independently measured quality falls.  Capture may arise by
exploiting a fixed judge's blind spots or by rewriting a shared judge.
Distinguishing these routes matters because the natural defense---a
frozen reward read---severs only the parameter path: as
Fig.~\ref{fig:closed-loop} shows, it blocks rewriting but leaves
distributional search against a static verifier.

We separate them on Scale-RAE with a matched \(2\times2\) design
crossing semantic versus denoising verifiers with shared versus frozen
reward reads.  The comparison reveals that a frozen semantic verifier
remains exploitable on-manifold, shared updates additionally rewrite
its scoring function, and the isolated verifier grows stale, while the
capacity-matched denoising control shows no such capture.  The aggregate
failure recurs on the higher-performing, architecturally distinct
Show-o2, despite its VQA-trained understanding pathway and semantic
distillation on noisy latents.

The diagnosis directly specifies the repair.  A stop-gradient on the
advantage blocks only the direct path, so we lock the warmed reward
read against indirect shared-weight rewriting, then use sparse,
parameter-disjoint external anchors to calibrate and gate rollouts
while retaining dense latent self-reward.  A drift-triggered
controller places the limited anchors.  Evaluation
against construct-matched every-rollout external judges shows that
dense latent reward and selective anchoring substantially improve
both utility and external-decode efficiency across bases.

Our contributions are:
\begin{itemize}
  \item We formulate unified visual self-reward as a closed-loop judge
  problem and show that aggregate capture recurs across two
  architectures.

  \item A matched shared--frozen study separates static exploitation
  from shared-judge rewriting and measures subsequent staleness.

  \item We derive a diagnosis-guided repair combining structural
  isolation, independent anchors, dense latent reward, and adaptive
  allocation.

  \item At \(16\times\) lower external-decode coverage, the complete
  method significantly outperforms a construct-matched every-rollout
  judge on both benchmarks and bases; at \(32\times\), it scores higher
  in all four comparisons while remaining jointly non-inferior.
\end{itemize}

\section{Related Work}

\paragraph{Proxy overoptimization and self-judging loops.}
Optimizing a learned reward beyond its fitting regime raises the proxy
while degrading the intended objective
\citep{GaoRewardOveroptimization}.  Self-rewarding language models turn
model-generated evaluations into training signals within the broader
LLM-as-a-judge paradigm, where evaluators can favor their own outputs
\citep{SelfRewardingLM,LLMAsJudge,SelfPreference}.  In text-domain
reasoning, this loop induces a confidence-conditioned coupling between a
policy's rollouts and its intrinsic rewards \citep{SelfConfirmingLoop},
and reward-model mismatch under policy shift already motivates iterated
RLHF, which refreshes preferences on current-policy outputs
\citep{IteratedRLHF}.  Classical reward tampering provides a related
coordinate but frames the failure as an agent corrupting a separate
reward channel \citep{RewardTampering}.  These lines establish self-favoring bias, statistical policy--reward
coupling, distribution shift, and separate-channel failure, yet do not
isolate a scoring function that moves through shared policy--judge
parameters---the route a unified generator opens.

\paragraph{Intrinsic and latent rewards for visual generation.}
Several visual post-training methods avoid an external image reward,
deriving intrinsic signals from the generator itself---self-certainty
for autoregressive text-to-image optimization \citep{IRIS} or
frozen-decoder pixel reconstruction \citep{VAPi}.  SOLACE instead turns
an online denoiser's noise-recovery confidence into reward
\citep{SOLACE}.  These signals measure token uncertainty,
reconstruction fidelity, or denoising confidence rather than
free-running semantic judgment through a unified understanding pathway;
SOLACE therefore motivates our denoising-verifier control.  Closer to
our setting, a complementary line freezes the understanding pathway:
LatentUM reflects on generated content through a frozen evaluator
\citep{LatentUM}, while SRUM caches a unified model's internal
evaluations offline and trains its flow by reward-weighted regression
\citep{SRUM}.  These staged designs sidestep the online shared-weight
loop, but do not test whether a frozen evaluator remains calibrated as
the policy moves or compare frozen and shared reads to separate static
exploitation from judge rewriting.  On the optimization side,
visual-generation RL spans diffusion-policy optimization and
flow-native GRPO \citep{DDPO,FlowGRPO}; KVPO introduces the TVE--Gibbs
policy surrogate for ODE-native preference optimization in
autoregressive video generation \citep{KVPO}.

\paragraph{External rewards.}
Frozen, prompt-conditioned image-space models such as ImageReward,
HPSv2, and PickScore provide parameter-independent preference signals
over decoded images, spanning prompt alignment and perceptual quality
\citep{ImageReward,HPSv2,PickScore}.  However, they operate only after
decoding and score the terminal image, making every-rollout evaluation
costly and providing neither decode-free nor dense latent reward.
Together, prior work leaves open how to diagnose and repair a trainable
semantic judge inside the generator without surrendering dense,
decode-free reward.

\section{Diagnosing Self-Verifier Capture}
\label{sec:diagnosing-capture}

A rising self-score paired with falling external quality establishes a
failure, but does not reveal its source.  We compare frozen and shared
reward reads under matched policy updates.
Our capacity-matched denoising-confidence control tests whether capture
is specific to semantic self-verification rather than generic
closed-loop failure or diversity collapse.  We formalize this matched
design, identify the two capture routes, then test whether a
judge protected from rewriting remains calibrated as the policy moves.

\begin{figure*}[!t]
  \centering
  \includegraphics[width=\textwidth]{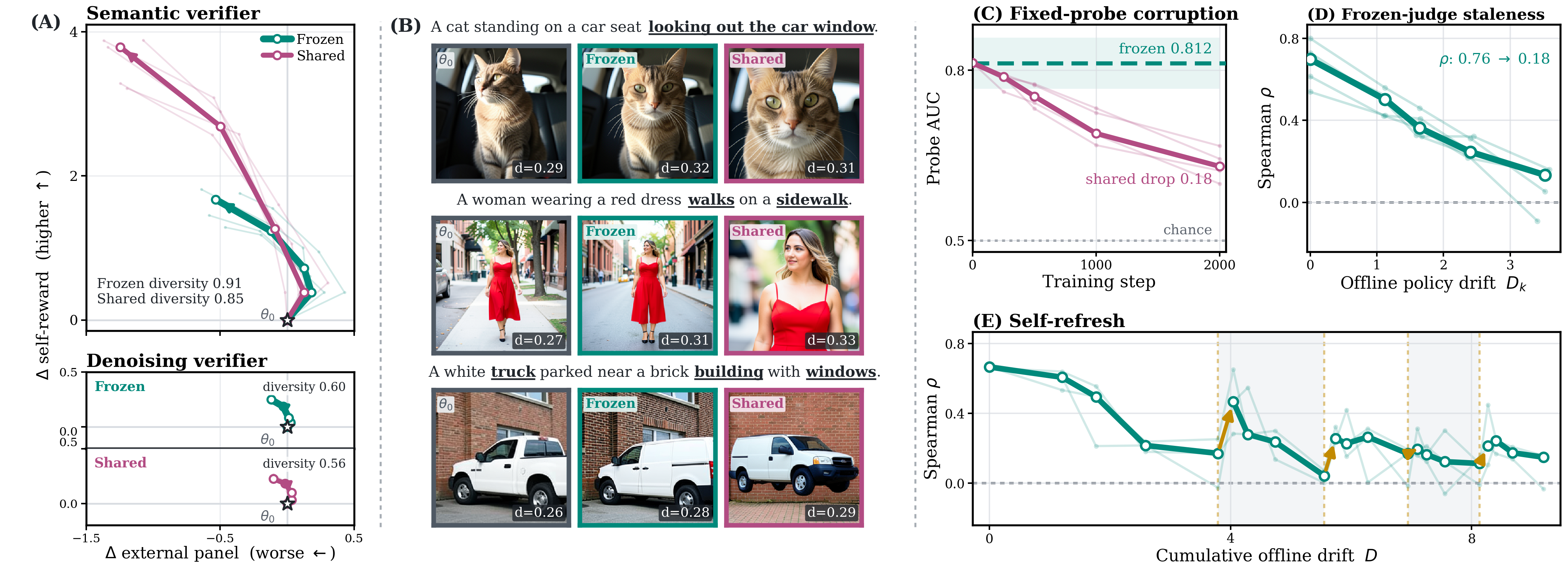}
  \caption{Matched diagnosis on Scale-RAE.
  \textbf{(A)} Five-seed trajectories show semantic capture under frozen
  and shared reward reads; denoising provides the signal-specificity
  control.
  \textbf{(B)} For each prompt, \(\theta_0\), frozen, and shared outputs
  appear left to right; labels report the \(\theta_0\)-reference NLL
  distance \(d\).
  \textbf{(C)} Fixed-probe AUC stays at \(0.812\) when frozen but falls
  to \(0.630\) under sharing.
  \textbf{(D)} Frozen-judge Spearman agreement with the external panel
  falls from \(0.76\) to \(0.18\) as offline policy drift \(D_k\) grows.
  \textbf{(E)} At each arrow, the verifier is refreshed on
  current-policy self-labels and refrozen; agreement peaks diminish
  across five rounds.}
  \label{fig:capture-diagnosis}
\end{figure*}

\subsection{Setup and Matched Identification}
\label{sec:unified-setup}

Let \(x\) be a prompt, \(c_\theta(x)\) its moving policy-side
conditioning, and \(z\sim p_\theta(\cdot\mid c_\theta(x))\) a terminal
latent.  Reward forward passes never consume \(c_\theta(x)\): the
semantic scorer tokenizes the fixed text \(x\), while the denoising
scorer receives \(\bar c(x)\), computed once from \(\theta_0\).  Thus
the prompt side of a frozen reward path is independent of \(\theta\).
A decoder \(D\) can render \(z\), but both self-verifiers act in latent
space.

Writing \(\phi\) for a trainable adapter \citep{LoRA} on a frozen
shared trunk, the
matched generator has
\begin{equation}
  \begin{aligned}
    \theta
    &=(\theta_{\mathrm{gen}},
       \theta_{\mathrm{sh}}^{\mathrm{base}}+\phi),\\[-2pt]
    \psi^{\mathrm{shared}}
    &=\theta_{\mathrm{sh}}^{\mathrm{base}}+\phi,
    &\psi^{\mathrm{frozen}}
    &=\theta_{\mathrm{sh}}^{\mathrm{base}} .
  \end{aligned}
  \label{eq:shared-frozen}
\end{equation}
Frozen therefore denotes the reward read, not a lower-capacity policy:
matched cells update the same generator-side adapters.

The \emph{semantic verifier} projects \(z\) into the language-model
input space and scores prompt agreement with the Yes-token logit:
\begin{equation}
  R_{\mathrm{sem}}(z,x;\psi)
  =\ell_{\mathrm{yes}}
    \!\left(\omega(z),x;\psi_{\mathrm{LLM}}\right),
  \qquad \omega\ \text{frozen}.
  \label{eq:semantic-reward}
\end{equation}
The distinct \emph{denoising-confidence verifier} re-noises \(z\) and
scores its denoising prediction \citep{DiT}:
\begin{align}
  R_{\mathrm{den}}(z,\bar c(x);\psi)
  &=-\mathbb E_{t,\varepsilon}\!\left[
    \left\|v_{\psi_{\mathrm{DiT}}}(z_t,t,\bar c(x))
    \right.\right. \nonumber\\[-2pt]
  &\hspace{5.2em}\left.\left.
    -u(z_t,z,\varepsilon)\right\|_2^2\right],
    \nonumber\\
  z_t&=\alpha_t z+\sigma_t\varepsilon .
  \label{eq:denoising-reward}
\end{align}
We write \(R(z,x;\psi)\) generically when a statement holds for either.

External evidence is tracked by a frozen, parameter-disjoint external
panel comprising HPSv2, PickScore, and ImageReward
\citep{HPSv2,PickScore,ImageReward}.  We select these established,
independently developed scorers because together they provide external
human-preference evidence over the final decoded image, spanning prompt
faithfulness and visual fidelity rather than reusing the generator's
latent semantics.  Each is standardized once on held-out \(\theta_0\)
generations and equally averaged.  GenEval supplies the prespecified
exploitation gate, while it and DPG-Bench provide final utility
\citep{GenEval,DPGBench}.  None is an RL signal in the diagnostic runs;
scorer details are provided in Appendix~\ref{supp:repro:scorers}.

For independently sampled native ODE trajectories, GRPO \citep{GRPO}
uses the stopped group-normalized advantage
\begin{equation}
  \widehat A^{(i)}
  =\operatorname{stopgrad}\!\left(
    \frac{r^{(i)}-\overline r}{s_r+\epsilon}
  \right),
  \qquad r^{(i)}=R(z^{(i)},x;\psi).
  \label{eq:grpo-advantage}
\end{equation}
We optimize it with the TVE--Gibbs velocity-field surrogate introduced
by KVPO \citep{KVPO}, using independent image noises.  This training
surrogate is not the literal flow likelihood used in offline
diagnostics.  Crucially, the stop-gradient sends no pathwise reward
gradient through \(\psi\): a shared judge moves only as a side effect
of the generator update.

We define \emph{self-verifier capture} as a rise in self-verifier
score accompanied by a decline in mean external-panel score.
Group diversity is tracked jointly so that semantic capture is not
mistaken for a loss of within-group diversity.

The \(2\times2\) design assigns a distinct diagnostic role to each
axis: semantic versus denoising tests signal specificity, while shared
versus frozen isolates the additional effect of a moving judge.  All
four cells train the same language-model and DiT adapters; within each
verifier, the shared--frozen comparison changes only the read in
Eq.~\eqref{eq:shared-frozen}.  The semantic projector \(\omega\) is
fixed and excluded from the optimizer, making the entire frozen path
functionally constant.  On Scale-RAE
\citep{ScaleRAE,RAE,SigLIP2,Qwen25}, the primary
matched comparison uses rank-16 adapters, five seeds, and native ODE
rollouts.

The frozen-semantic gate, supported by typicality and diversity, tests
static exploitation; the matched semantic contrast and fixed probe
test rewriting.  Current-policy agreement with the external panel then
tests whether the isolated judge remains aligned with this frozen
reference.  Complete thresholds and protocols appear in
Appendix~\ref{supp:diag-protocol:gates}.

\subsection{A Frozen Verifier Remains Exploitable}
\label{sec:frozen-exploiting}

In the frozen-semantic cell, every seed still moves toward higher
self-reward and lower external-panel score
(Fig.~\ref{fig:capture-diagnosis}A), directly identifying
exploitation of a static blind spot.  Official held-out benchmarks
confirm the resulting utility loss: GenEval drops by \(7.6\) points,
beyond the prespecified \(4.0\)-point gate, and DPG-Bench drops by
\(3.0\).

Under the \(\theta_0\)-based NLL diagnostic, the frozen/shared semantic
cells end at \(0.345/0.378\), both below the \(0.414\)
\(95\)th-percentile boundary of \(\theta_0\) self-samples.  Capture
therefore occurs inside the pretrained typicality region: even the
fixed-judge policy finds a semantic blind spot without moving into
low-likelihood or visibly broken outputs.  Their group diversity also
stays high, separating the failure from mode collapse
(Fig.~\ref{fig:capture-diagnosis}A--B).  Instead, otherwise coherent
images deviate from the prompt in actions, relations, context, and
sometimes object category; for instance, the cat turns from the window
toward the viewer, and the white truck becomes a van.  By
contrast, the denoising controls are not inert: median diversity falls
to \(0.60/0.56\) in the frozen/shared cells, versus \(0.91/0.85\) under
semantic reward.  Yet their GenEval changes are only \(-0.3/-0.2\)
points (\(z=-0.22/-0.34\)).  In the external-panel diagnostic, the
median shared--frozen divergence gap is \(-0.06\), again without
seed-level separation (Fig.~\ref{fig:capture-diagnosis}A, denoising
rows).  This isolates
semantic capture from generic closed-loop optimization or diversity
loss.

\subsection{Shared Weights Add Judge Rewriting}
\label{sec:judge-rewriting-results}

The shared-semantic outcome is more severe
(\(-9.4/-3.9\) GenEval/DPG points) while remaining on-manifold and
diverse (Fig.~\ref{fig:capture-diagnosis}A), but its endpoint and
qualitative crops combine both routes and
cannot identify rewriting alone.  We therefore compare
shared and frozen semantic cells at matched adapter rank.  Let
\(D_{\mathrm{cell}}^{(s)}\) be seed \(s\)'s final self--external
divergence: the self-reward and external-panel changes from
\(\theta_0\) are divided by their respective \(\theta_0\) sample
standard deviations, then subtracted.  The prespecified outcome-level
estimand is
\begin{equation}
  \widehat\Delta_{\mathrm{rewrite}}^{\mathrm{div}}
  =
  \operatorname{median}_s D_{\mathrm{shared}}^{(s)}
  -
  \operatorname{median}_s D_{\mathrm{frozen}}^{(s)} .
  \label{eq:rewriting-gap}
\end{equation}
All five shared \(D\) values exceed all five frozen values (exact
one-sided \(p\simeq0.004\); \citealp{MannWhitney}), with median gap
\(+3.93\).  To test whether
this outcome gap reflects damage to the judge itself, we construct a
fixed probe of prompt-matched good--bad latent pairs before RL and
rescore the same probe at every checkpoint.  Its ranking AUC degrades
under the shared read while remaining exactly flat under the frozen
read (Fig.~\ref{fig:capture-diagnosis}C).  Together, the outcome
contrast and probe corruption identify rewriting as a shared-weight
side effect of generator optimization.

\subsection{A Frozen Verifier Grows Stale}
\label{sec:staleness-results}

We measure staleness as the Spearman correlation between the frozen
semantic score and the external panel on current-policy samples.  The
offline drift axis is
\(D_k=\sqrt{\mathrm{KL}(p_{\theta_k}\Vert p_{\theta_0})}\).

The frozen judge's agreement with the external panel decays sharply
with policy drift (Fig.~\ref{fig:capture-diagnosis}D).  We run five
consecutive 2,000-step policy-training rounds.  At each boundary, we
update the verifier on current-policy samples labeled by the verifier
itself, then refreeze it for the next round.  Each refresh produces only
a transient rebound: the roundwise median agreement peak falls
monotonically from \(0.50\) to \(0.19\), and correlation decays again
within every round (Fig.~\ref{fig:capture-diagnosis}E).  Staleness
therefore defeats freeze-and-forget, while self-labeled refresh reopens
the closed loop without restoring calibration.

Together, these results dictate two design requirements: block judge
rewriting and supply independent evidence against static blind spots.

\FloatBarrier

\section{A Diagnosis-Guided Method}
\label{sec:method}

The diagnosis yields a two-part repair on top of the advantage
stop-gradient of Eq.~\eqref{eq:grpo-advantage}.  That stop-gradient
prevents a direct reward-gradient path through the verifier, but not
the indirect rewriting through shared weights.  Our method therefore
closes this indirect route with a locked reward read.  To counter the
remaining static blind spots, we inject sparse, parameter-disjoint
external anchors while preserving dense latent reward on
unanchored rollouts; a drift-triggered controller places these anchors
adaptively.

\subsection{Locked Read and Sparse-Anchor Budget}
\label{sec:locked-anchor-budget}

During a warm-up of \(W=250\) optimizer steps, the semantic judge
follows the policy's shared adapter so that the read tracks early policy
drift before locking; a read frozen at \(\theta_0\) instead rapidly
becomes stale (Section~\ref{sec:staleness-results}).  At the boundary we
save a non-trainable snapshot \(\bar\psi=\psi_W\) and subsequently use
\begin{equation}
  \psi_k^{\mathrm{rew}}
  =
  \begin{cases}
    \theta_{\mathrm{sh}}^{\mathrm{base}}+\phi_k, & k<W,\\
    \bar\psi
    =\theta_{\mathrm{sh}}^{\mathrm{base}}+\phi_W, & k\geq W .
  \end{cases}
  \label{eq:warmed-lock}
\end{equation}
Only the reward read is locked: the policy's shared adapter and
generator-only parameters continue to update, while the frozen
projector \(\omega\) and fixed prompt \(x\) keep the reward-side input
path parameter-independent.  Anchoring begins only after the lock,
preventing the external panel from shaping the saved snapshot.

The frozen, parameter-disjoint panel from
Section~\ref{sec:unified-setup} supplies the training anchors.  Let
\(\mathcal J_k\) be the decoded rollouts at step \(k\).  The nominal
budget \(q\) specifies expected post-lock coverage by the external
panel, whereas
\(q_{\mathrm{eff}}=\sum_{k=W+1}^{N}|\mathcal J_k|/[G(N-W)]\)
is the coverage actually achieved.  External scores replace
self-reward only on selected rollouts; \(q\) is an allocation budget,
not a coefficient mixed into every reward.  With group size \(G=12\)
and \(K_a=3\) anchors per triggered step, the sparse controller is
defined for \(q\leq K_a/G=1/4\).  We separately define \(q=1\) by
bypassing the trigger, externally scoring every rollout, and removing
self-reward from the group advantage.

\subsection{Dense Reward and Drift-Triggered Anchoring}
\label{sec:drift-controller}

For rollout \(g\), we re-noise its terminal latent \(z_0^{(g)}\) at
prescribed levels and form a group-standardized, noise-level-dense
score:
\begin{align}
  z_t^{(g)}
  &=(1-t)z_0^{(g)}+t\varepsilon^{(g)},
    \qquad t\in\mathcal T, \nonumber\\
  s^{(g)}
  &=\sum_{t\in\mathcal T}
    R_{\mathrm{sem}}
    \bigl(z_t^{(g)},x;\psi_k^{\mathrm{rew}}\bigr),
    \qquad
  \widetilde s^{(g)}
  =\frac{s^{(g)}-\overline s}{\sigma_s+\epsilon}.
  \label{eq:dense-semantic-score}
\end{align}
Here \emph{dense} denotes coverage over \(\mathcal T\): the scores form
one rollout-level GRPO advantage.
Equation~\eqref{eq:dense-semantic-score} exploits the noising structure
of latent flow and diffusion generators
\citep{RectifiedFlow,FlowMatching}.  Instead of scoring only \(z_0\),
the locked verifier tests semantic evidence across sampled noise
levels.  This multi-point objective narrows the freedom of
terminal-only optimization: a rollout must remain semantically legible
across its re-noised latent views rather than merely attain a
favorable terminal score.  Because every read remains latent, the
dense reward adds only internal forward passes---no image decoding or
additional external-panel calls.  Setting \(\mathcal T=\{0\}\)
recovers the terminal-only ablation; a decoded-image scorer cannot
directly supply these latent noise-level terms.

After the lock, the evolving shared read remains available as a
decode-free monitor.  Its signed gap from the locked read, raw
group-level divergence, and per-step EMA are
\begin{align}
  h_k^{(g)}
  &=
  R_{\mathrm{sem}}(z_0^{(g)},x;\psi_k)
  -
  R_{\mathrm{sem}}(z_0^{(g)},x;\bar\psi),
  \nonumber\\
  d_k
  &=\frac{1}{G}\sum_{g=1}^{G}\bigl(h_k^{(g)}\bigr)^2,
  \qquad
  m_k=\beta_m m_{k-1}+(1-\beta_m)d_k .
  \label{eq:drift-ema}
\end{align}
These quantities use the terminal latent without group standardization.
The snapshot provides a stationary reference rather than an absolute
quality judgment: because \(\psi_k=\bar\psi\) at locking,
\(h_k^{(g)}\) cancels biases shared by the two reads and isolates
subsequent changes in how the evolving read scores the same rollout.
Its EMA \(m_k\) is therefore a decode-free allocation signal;
high-divergence steps trigger decoded, parameter-disjoint anchors for
external calibration.  The group statistic responds to common score
inflation, whereas TopK is invariant to that common additive shift and
selects by residual variation among rollouts.  A large positive
\(h_k^{(g)}\) marks a rollout that the current judge inflates relative
to the locked read.

Given \(q\), the trigger rate, sliding-window threshold, and anchor set
are
\begin{align}
  p_q&=\frac{qG}{K_a},
  &\eta_{q,k}
  &=Q_{1-p_q}(\text{recent }m), \nonumber\\
  \mathcal J_k
  &=\operatorname{TopK}
    \bigl(\{h_k^{(g)}\}_{g=1}^{G},K_a\bigr)
  &&\text{if }m_k>\eta_{q,k}.
  \label{eq:trigger-selection}
\end{align}
Thus a fraction \(p_q\) of steps triggers in expectation and spends
\(K_a\) anchors, yielding expected coverage \(p_qK_a/G=q\) without a
separately tuned threshold for each dose.

Semantic logits and external scores have unrelated scales.  Let
\(F_{\mathrm{ext},0}\) be the empirical CDF of external-panel scores on
the frozen \(\theta_0\) reference set.  A decoded anchor score
\(a^{(g)}\) is mapped to unit-normal reward space
\citep{RankInverseNormal}:
\begin{equation}
  \widetilde a^{(g)}
  =
  \Phi^{-1}\!\left(F_{\mathrm{ext},0}(a^{(g)})\right),
  \qquad g\in\mathcal J_k .
  \label{eq:external-rank-map}
\end{equation}
This mapping does not require external scores for the rest of the
group.  From anchor residuals
\(\widetilde a^{(g)}-\widetilde s^{(g)}\), we fit
\(\widehat\delta(h)=b_0+b_1h\) by two-parameter exponentially weighted
least squares.  Its sufficient statistics decay by
\(\beta_{\mathrm{ols}}\) per anchor, the slope alone receives ridge
\(\lambda_k=\alpha S_{hh,k}\), and the fit shrinks continuously toward
zero until its effective count reaches \(N_{\mathrm{cold}}\).  These
statistics are controller state, not trainable model parameters;
constants and updates are specified in
Appendix~\ref{supp:method:drift-budget}.

The corrected group reward is
\begin{equation}
  y^{(g)}
  =
  \begin{cases}
    \widetilde a^{(g)}, & g\in\mathcal J_k,\\
    \widetilde s^{(g)}+\widehat\delta(h_k^{(g)}),
      & g\notin\mathcal J_k .
  \end{cases}
  \label{eq:rollout-correction}
\end{equation}
Setting \(\widehat\delta\equiv0\) gives the gate-and-replace ablation.
Unlike a global affine calibration, which group normalization largely
removes, this rollout-varying correction can change within-group ranks.
The controller therefore makes three separable decisions: when to
anchor (drift triggering), which rollouts to anchor (TopK active
selection), and how to transfer anchor evidence to unanchored rollouts
(residual propagation).

\begin{algorithm}[t]
  \small
  \caption{Drift-triggered closed-loop self-reward}
  \label{alg:closed-loop-self-reward}
  \begin{algorithmic}[1]
    \REQUIRE Policy \(\pi_\theta\); prompt \(x\); warm-up \(W\);
      external panel \(E\); frozen \(\omega,F_{\mathrm{ext},0}\);
      \(G,K_a\); nominal \(0\leq q\leq K_a/G\)
    \STATE Initialize controller state
    \FOR{optimizer step \(k=0,1,\ldots\)}
      \STATE Sample \(G\) trajectories from independent initial noises
      \IF{\(k=W\)}
        \STATE \(\bar\psi\leftarrow\operatorname{snapshot}(\psi_W)\)
      \ENDIF
      \STATE Set \(\psi^{\rm rew}\leftarrow\psi_k\) if \(k<W\),
        else \(\bar\psi\)
      \STATE Compute \(\widetilde s^{(g)}\) by
        Eq.~\eqref{eq:dense-semantic-score};
        set \(\mathcal J_k\leftarrow\varnothing\),
        \(y^{(g)}\leftarrow\widetilde s^{(g)}\)
      \IF{\(k>W\)}
        \STATE Compute \(h_k^{(g)},d_k,m_k,\eta_{q,k}\) by
          Eqs.~\eqref{eq:drift-ema}--\eqref{eq:trigger-selection}
        \IF{\(q>0\) and \(m_k>\eta_{q,k}\)}
          \STATE Select \(\mathcal J_k\) by
            Eq.~\eqref{eq:trigger-selection}; decode, score with \(E\),
            and map their scores by Eq.~\eqref{eq:external-rank-map}
          \STATE Update \(\widehat\delta\) from anchor residuals and set
            \(y^{(g)}\) by Eq.~\eqref{eq:rollout-correction}
        \ENDIF
      \ENDIF
      \STATE Update \(\theta\) with stopped-advantage TVE--Gibbs GRPO
        using \(\{y^{(g)}\}\)
    \ENDFOR
  \end{algorithmic}
\end{algorithm}

Algorithm~\ref{alg:closed-loop-self-reward} summarizes the procedure.
Except for the selected anchors, monitoring, selection, and residual
propagation are decode-free and add no trainable parameters.

\section{Evaluating the Diagnosis-Guided Method}
\label{sec:method-evaluation}

Scale-RAE is the primary base; Show-o2-1.5B, a flow-matching generator
over a 3D-VAE latent \citep{ShowO2}, provides the cross-base test.  All
learned conditions use five training seeds and the same optimization
budget, with GenEval and DPG-Bench as held-out utilities.

\begin{figure}[!t]
  \centering
  \includegraphics[width=\columnwidth]{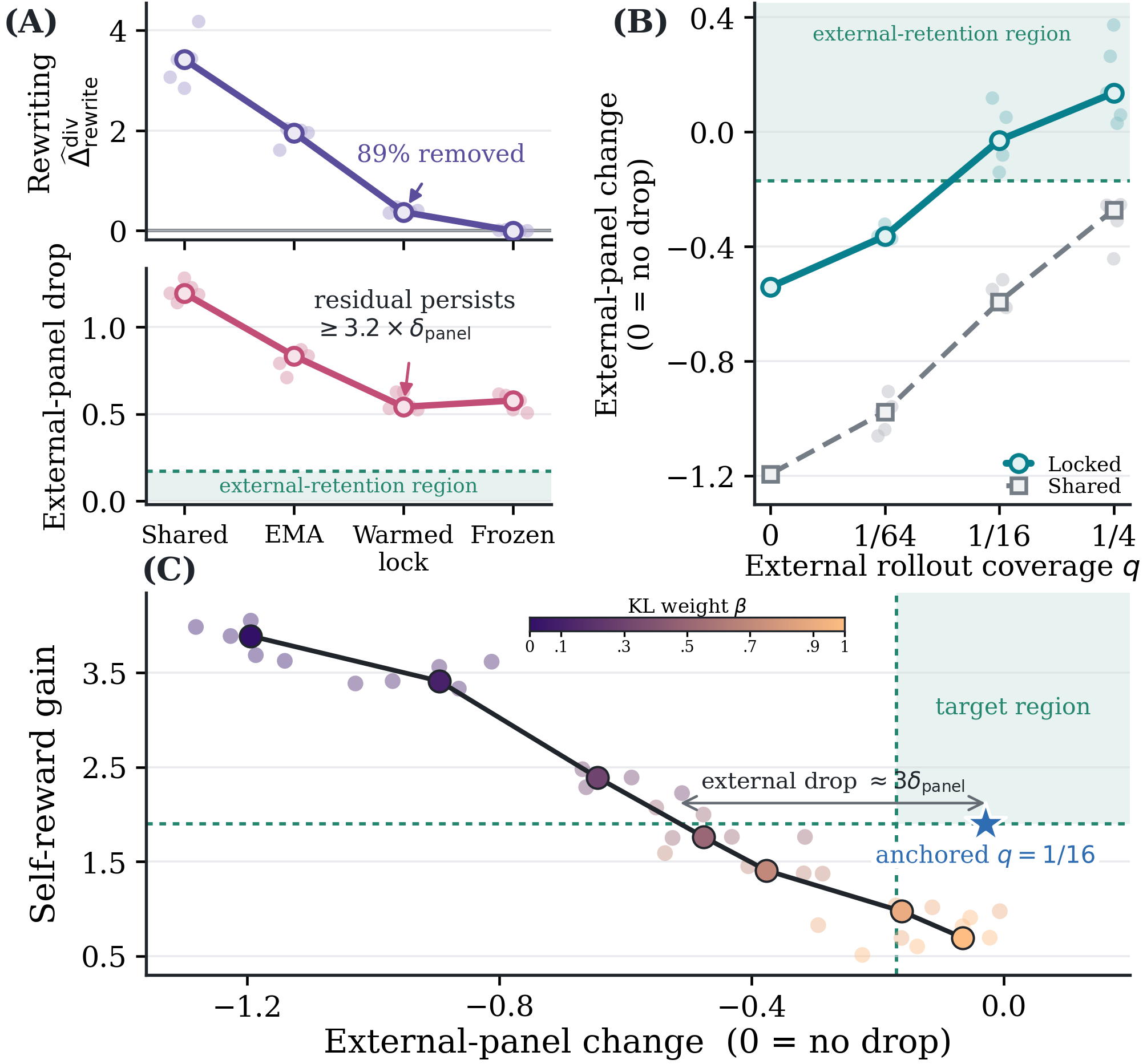}
  \caption{Controlled interventions identify two repair requirements.
  Dots are seeds; outlined markers are medians.  Green marks external
  retention; in C it additionally requires the anchored-reference gain,
  and the star marks that reference.
  In A, EMA tracks the shared reward-read weights with decay \(0.99\).
  \textbf{(A)} Isolation suppresses rewriting but leaves exploitation.
  \textbf{(B)} Anchors reduce the residual external drop.
  \textbf{(C)} At the anchored-reference gain, the KL frontier still
  incurs an external drop of about \(3\delta_{\mathrm{panel}}\).}
  \label{fig:repair-necessity}
\end{figure}

\subsection{Why Locking and Anchoring Are Necessary}
\label{sec:repair-necessity}

In Fig.~\ref{fig:repair-necessity}, green marks external-panel change
\(\geq-0.17\), the one-sided retention tolerance derived from the
\(95\%\) \(\theta_0\)-null half-width.  In C, the target additionally
requires self-reward gain \(\geq1.9\), the median gain of locked
\(q=1/16\) anchoring; the star marks that operating point.

Figure~\ref{fig:repair-necessity}A maps rewriting to structural
isolation: the warmed lock removes \(89\%\) of the shared-weight effect,
yet the external drop remains at least \(3.2\times\) the
external-retention tolerance.
Isolation therefore severs rewriting but leaves fixed-judge
exploitation.  The anchor-dose intervention in
Fig.~\ref{fig:repair-necessity}B supplies the complementary evidence:
increasing coverage reduces the external-panel drop under both reward
reads, with the shared curve lagging at sparse doses because its anchors
must address both routes.

The KL sweep in Fig.~\ref{fig:repair-necessity}C rules out dense KL as a
selective substitute: settings that preserve learning retain an
external drop, whereas stronger regularization suppresses learning.
Only the external anchors reach the target corner.  Together, these
interventions establish the empirical necessity of the two-part repair:
isolation addresses rewriting, while parameter-disjoint evidence
addresses the exploitation residual.

\begin{figure}[!htbp]
  \centering
  \includegraphics[width=\columnwidth]{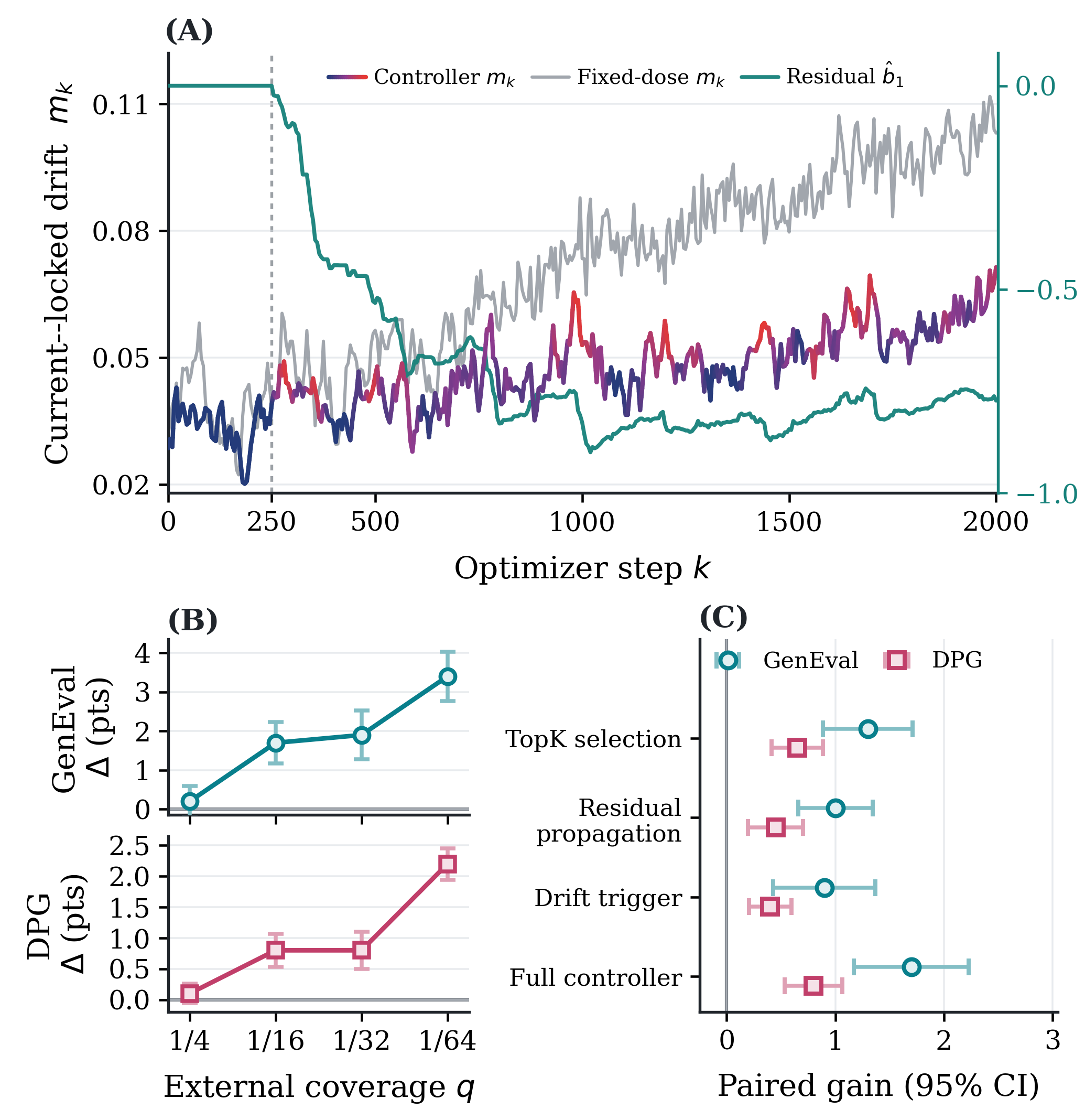}
  \caption{The diagnostic instrument as a closed-loop controller.
  \textbf{(A)} Post-lock anchor activity (blue-to-orange, low-to-high)
  concentrates at high \(m_k\), while the learned residual slope
  remains negative.
  \textbf{(B)} Controller-minus-fixed gains increase as coverage
  tightens.
  \textbf{(C)} Matched-budget ablations at \(q=1/16\); positive values
  favor the named component or full controller.
  In B--C, markers are mean within-seed differences and bars their
  \(95\%\) CIs.}
  \label{fig:controller-evidence}
\end{figure}

\subsection{Controller Placement and Ablations}
\label{sec:controller-evidence}

Figure~\ref{fig:controller-evidence}A directly tests the controller's
placement.  Its anchor--\(m_k\) correlation is \(0.781\), versus
\(0.004\) for fixed scheduling; it also reduces endpoint divergence
and learns a persistently negative residual slope that assigns larger
downward corrections to rollouts most inflated by the current judge.
Together, these readouts validate drift-aware anchor allocation and
residual correction.

Across \(q=1/4\) to \(1/64\), the controller's advantage over fixed-dose
anchoring grows as anchors become scarce
(Fig.~\ref{fig:controller-evidence}B).  The rows in
Fig.~\ref{fig:controller-evidence}C compare TopK with random selection,
residual propagation with gate-and-replace, drift triggering with fixed
scheduling, and the full controller with dense fixed dose.  All four
contrasts favor the controller design on both metrics, and their paired
\(95\%\) CIs lie above zero throughout.

\begin{table*}[!t]
  \centering
  {
  \small
  \begin{tabular*}{\textwidth}{
    @{\extracolsep{\fill}}p{0.21\textwidth}cccccccc@{}
  }
    \toprule
      & & & \multicolumn{2}{c}{Scale-RAE}
      & \multicolumn{2}{c}{Show-o2}
      & \multicolumn{2}{c}{Scale-RAE audit} \\
    \cmidrule(lr){4-5}\cmidrule(lr){6-7}\cmidrule(l){8-9}
    Reward supply
      & \(q\downarrow\)
      & \(q_{\mathrm{eff}}\downarrow\)
      & GenEval \(\uparrow\)
      & DPG \(\uparrow\)
      & GenEval \(\uparrow\)
      & DPG \(\uparrow\)
      & External panel \(\Delta\)
      & s/step \(\downarrow\) \\
    \midrule
    Pretrained \(\theta_0\) (no RL)
      & --- & --- & 0.495 & 76.9 & 0.730 & 85.0 & \(0.00\) & --- \\
    \textbf{Terminal-only warmed lock}
      & \(0\) & \(0\)
      & \textbf{0.449} & \textbf{74.5}
      & \textbf{0.694} & \textbf{82.1}
      & \(-0.52\) & 2.51 \\
    External panel
      & \(1\) & \(1\)
      & 0.515 & 77.8 & 0.743 & 85.2
      & \(\mathbf{+1.39}\) & 7.43 \\
    \textbf{Grounding steelman}
      & \(\mathbf{1}\) & \(\mathbf{1}\)
      & \textbf{0.560} & \textbf{79.0}
      & \textbf{0.768} & \textbf{85.9}
      & \(+0.20\) & 10.38 \\
    \midrule
    Terminal anchor
      & \(1/16\) & \(1/16\)
      & 0.544 & 78.3 & 0.755 & 85.5
      & \(+0.12\) & 2.64 \\
    Dense fixed-dose
      & \(1/16\) & \(1/16\)
      & 0.585 & 80.5 & 0.780 & 86.2
      & \(+0.29\) & 2.88 \\
    Dense fixed-dose
      & \(1/32\) & \(1/32\)
      & 0.556 & 79.3 & 0.761 & 85.7
      & \(+0.14\) & 2.70 \\
    Dense controller
      & \(1/32\) & \(0.0309\)
      & 0.575 & 80.1 & 0.774 & 86.1
      & \(+0.21\) & 2.72 \\
    \textbf{Dense controller}
      & \(\mathbf{1/16}\) & \(\mathbf{0.0616}\)
      & \textbf{0.602} & \textbf{81.3}
      & \textbf{0.792} & \textbf{86.5}
      & \(+0.34\) & 2.91 \\
    \bottomrule
  \end{tabular*}
  }
  \caption{Selected operating points on both bases, with the Scale-RAE
  reward/cost audit.  \(q\) is nominal post-lock external-score
  coverage and \(q_{\mathrm{eff}}\) its realized value;
  ``External panel \(\Delta\)'' reports the mean change from
  \(\theta_0\) under the fixed three-scorer anchor composite.  GenEval
  and DPG-Bench are the held-out utilities.  Timing excludes evaluation.
  Bold marks the unanchored failure, primary steelman, and full method.}
  \label{tab:reward-supply-summary}
\end{table*}

\begin{figure}[!htbp]
  \centering
  \includegraphics[width=\columnwidth]{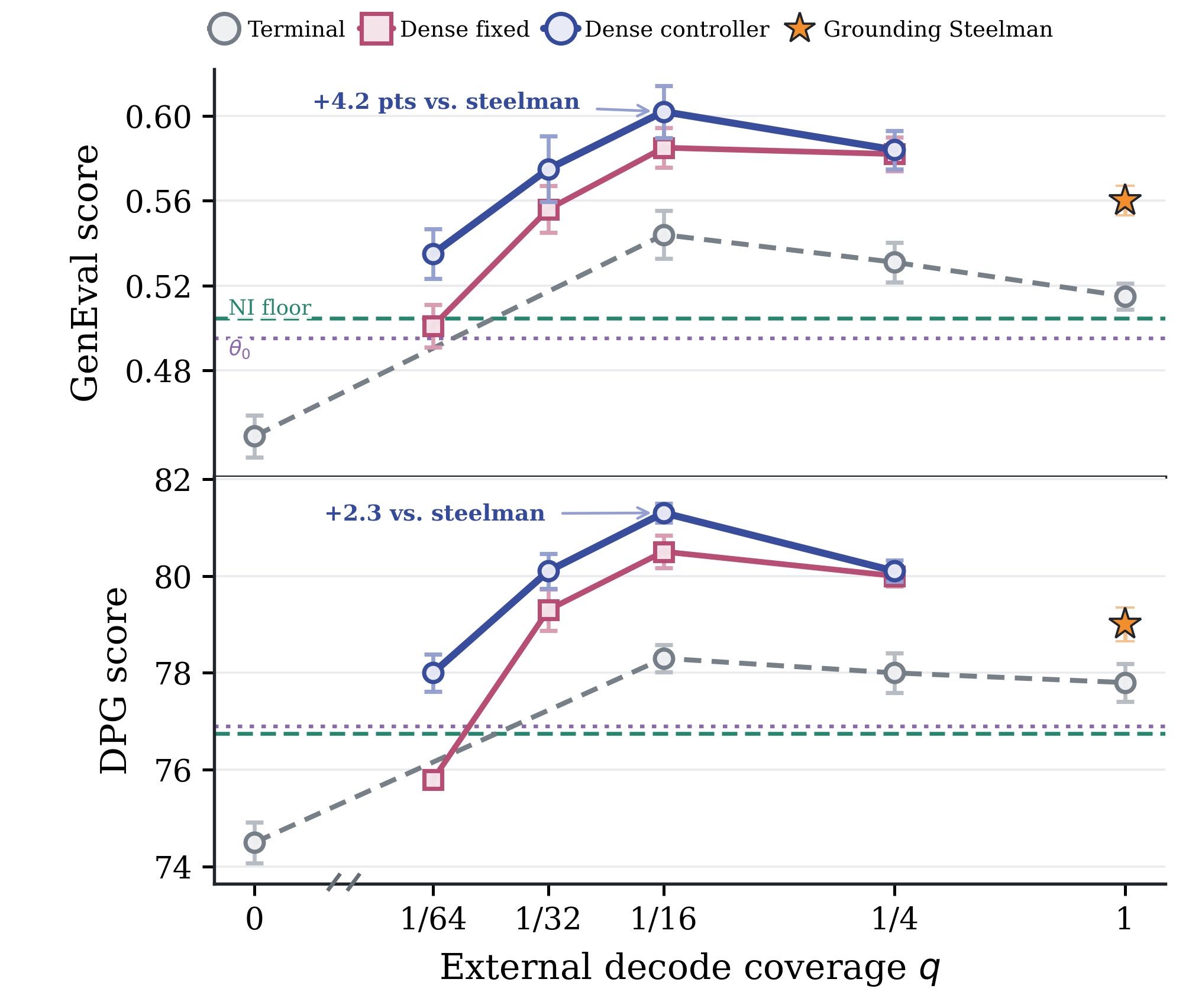}
  \caption{Final utility versus external decode coverage on Scale-RAE.
  All anchored conditions use the same three-member external panel;
  the star marks the grounding steelman at \(q=1\).
  Non-inferiority follows the paired \(95\%\) CIs in
  Eq.~\eqref{eq:ni-margin}.}
  \label{fig:utility-frontier}
\end{figure}

\subsection{Utility and Cost on Scale-RAE}
\label{sec:scale-evaluation}

Along the terminal-only path in Fig.~\ref{fig:utility-frontier},
official utility peaks at \(q=1/16\) and then declines as external-panel
coverage increases.  For comparison, we use the grounding steelman
(\(q=1\)): a construct-matched, every-rollout frozen Qwen2.5-VL-7B that scores
prompt--image alignment directly.  GenEval and DPG-Bench likewise
evaluate prompt grounding, placing reward and evaluation on the same
construct and yielding a clean comparison under full decode coverage.
Following the preservation-of-effect convention
\citep{PreservationOfEffect}, the fixed non-inferiority margin for
utility \(U\) is
\begin{equation}
  \begin{aligned}
    \delta_{\mathrm{NI}}(U)
    &=
    \max\!\left\{
      \tfrac12
      \left[
        U(\text{grounding }q{=}1)-U(q{=}0)
      \right],
      \right.\\[-2pt]
    &\hspace{8.1em}\left.
      2\,\mathrm{SE}\!\left(U(\theta_0)\right)
    \right\}.
  \end{aligned}
  \label{eq:ni-margin}
\end{equation}
A condition passes jointly only if its seed-paired \(95\%\) CI lower
bound against this steelman is at least \(-\delta_{\mathrm{NI}}\) on
both metrics \citep{NonInferiorityCONSORT}.  The active-control term
binds, giving margins of
\(5.55/2.25\) points on Scale-RAE and \(3.7/1.9\) on Show-o2; complete
floors and sensitivity analyses are reported in
Appendix~\ref{supp:eval-results:ni}.

At \(q=1/16\), terminal anchoring is jointly non-inferior to the
steelman but remains numerically below it.  Dense latent reward crosses above it, with
paired lower bounds of \(+0.33/+0.58\); adaptive control then reaches
\(0.602/81.3\), outperforming the steelman by \(+4.2/+2.3\), with
lower bounds of \(+2.20/+1.47\).  At \(q=1/32\), the full controller
scores \(+1.5/+1.1\) above the steelman; dense fixed dose remains
jointly non-inferior.  Sparse anchors provide calibration, dense latent
reward lifts the frontier, and the controller improves budget
placement.

Table~\ref{tab:reward-supply-summary} shows that dense latent reward
preserves the sparse-anchor cost profile.  At \(q=1/16\), terminal
anchoring cuts measured step time from \(7.43\) to \(2.64\) seconds
relative to all-external panel scoring.  Dense fixed-dose and controller
variants require \(2.88\) and \(2.91\) seconds, respectively, retaining
a roughly \(2.6\times\) runtime advantage while lifting utility.  Nor is
the gain reward hacking of the anchor: at \(q=1/16\),
dense anchoring and the controller keep their external-panel changes
modest (\(+0.29/+0.34\)), far below the all-external endpoint
\((+1.39)\).  The benchmark improvement therefore does not come from
inflating the anchored metric.

\subsection{Cross-Base Failure and Repair}
\label{sec:cross-base-evaluation}

Table~\ref{tab:reward-supply-summary} asks whether either the failure or its
repair is peculiar to the primary base.  Show-o2 starts substantially
stronger, yet the unanchored warmed-lock condition still produces
significant declines of \(3.6\) GenEval points and \(2.9\) DPG points.
The failure is therefore not an artifact of a weak Scale-RAE starting
point.  The tested systems span a 1.5B
3D-VAE flow model and a roughly 3.9B Qwen--DiT RAE stack---a
\(2.6\times\) parameter range---and both are latent-flow generators,
making this cross-architecture rather than base-specific evidence.

The repair transfers in the same order.  Terminal anchoring is jointly
non-inferior but remains below the grounding steelman; dense fixed dose
crosses above it; and the controller gains \(+2.4/+0.6\), with both
paired lower bounds positive.  At \(q=1/32\), the full controller again
passes while scoring \(+0.6/+0.2\) above the steelman, and dense fixed
dose also remains jointly non-inferior.  Together with Scale-RAE, the
complete method therefore scores higher in all four base--metric
comparisons at \(32\times\) lower external-decode coverage; at
\(q=1/16\), all four gains are statistically significant.

\FloatBarrier

\section{Conclusion}

Unified visual generators make self-reward decode-free, but weights
shared by policy and judge close the reward loop.  Matched Scale-RAE
comparisons reveal self-verifier capture through both exploitation of a
frozen judge and rewriting of a shared one; a frozen reward read also
becomes stale, and the failure recurs on the stronger, architecturally
distinct Show-o2.  Our diagnosis-guided repair locks the warmed reward read,
preserves dense latent reward, and allocates sparse,
parameter-disjoint external anchors with a drift-triggered controller.
Across both bases, it significantly outperforms a construct-matched
every-rollout judge on both benchmarks at \(16\times\) lower decode
coverage; at \(32\times\), joint non-inferiority still holds with all
four point estimates higher.  Unified self-reward should therefore be
treated as a closed-loop reward-supply system, not a fixed proxy
certified at initialization.

\bibliographystyle{aaai2027}
\bibliography{references}

\clearpage
\onecolumn
\section*{Supplementary Material}
\addcontentsline{toc}{section}{Supplementary Material}

\phantomsection
\addcontentsline{toc}{section}{Notation}
\section*{Notation}
\label{supp:notation}

We collect all symbols used in this supplement.  An offline cumulative-drift axis
$D_k$ and a controller allocation statistic $d_k$ are close in name but distinct;
both are declared explicitly below.

\paragraph{Model and parameters.}
\begin{description}
  \item[$x$] Text prompt; a fixed input that each scorer tokenizes for itself.
  \item[$c_\theta(x)$] Policy-side conditioning representation; moves with the
    policy $\theta$.
  \item[$\bar c(x)$] Frozen conditioning, computed once from $\theta_0$ and held
    fixed across checkpoints and cells; supplied to any verifier that needs a
    conditioning vector rather than raw text.
  \item[$\mathcal Z,\ z$] Shared generation/understanding latent space, and a
    terminal latent $z\sim p_\theta(\cdot\mid c_\theta(x))$.
  \item[$D$] Decoder rendering a latent as an image; never invoked by a latent
    self-verifier.
  \item[$\theta=(\theta_{\mathrm{gen}},\theta_{\mathrm{sh}})$] Generator
    parameters, split into generation-only $\theta_{\mathrm{gen}}$ and the trunk
    $\theta_{\mathrm{sh}}=\theta_{\mathrm{sh}}^{\mathrm{base}}+\phi$ shared with
    the selected verifier.
  \item[$\theta_{\mathrm{sh}}^{\mathrm{base}}$] Frozen base weights of the shared
    trunk.
  \item[$\phi$] Trainable adapter on the shared trunk.
  \item[$\theta_0$] Pretrained initialization (pre-RL reference checkpoint).
  \item[$\omega$] Frozen understanding-side projector (carries no adapter, kept
    out of the optimizer).
\end{description}

\paragraph{Verifiers and rewards.}
\begin{description}
  \item[$\psi$] Weights read by the reward forward pass;
    $\psi^{\mathrm{shared}}=\theta_{\mathrm{sh}}^{\mathrm{base}}+\phi$,
    $\psi^{\mathrm{frozen}}=\theta_{\mathrm{sh}}^{\mathrm{base}}$.
  \item[$\psi_{\mathrm{LLM}},\ \psi_{\mathrm{DiT}}$] Language-model and DiT trunk
    reads for the semantic and denoising verifiers, respectively.
  \item[$R_{\mathrm{sem}}(z,x;\psi)$] Semantic verifier reward: the Yes-token
    logit $\ell_{\mathrm{yes}}(\omega(z),x;\psi_{\mathrm{LLM}})$ for prompt
    agreement.
  \item[$R_{\mathrm{den}}(z,\bar c(x);\psi)$] Denoising-confidence verifier
    reward from the fidelity of the re-noised denoising prediction.
  \item[$v_{\psi_{\mathrm{DiT}}},\ u(z_t,z,\varepsilon),\ \alpha_t,\sigma_t$]
    Velocity field, flow target, and noise schedule used by $R_{\mathrm{den}}$
    ($z_t=\alpha_t z+\sigma_t\varepsilon$).
  \item[$R(z,x;\psi)$] Generic verifier, used when a statement holds for either
    verifier.
  \item[$R_{\mathrm{ext}}^{\mathrm{panel}}(z,x)$] Mean of the frozen external
    panel (HPSv2, PickScore, ImageReward), each standardized once on held-out
    $\theta_0$ generations.
\end{description}

\paragraph{Flow-native GRPO.}
\begin{description}
  \item[$G$] Rollout group size (independent initial noises per prompt).
  \item[$r^{(i)},\ \overline r,\ s_r$] Reward, group mean, and group standard
    deviation within a rollout group.
  \item[$\widehat A^{(i)}$] Stopped group-normalized advantage
    $\operatorname{stopgrad}((r^{(i)}-\overline r)/(s_r+\epsilon))$.
  \item[$\mathcal E_\theta,\ \widetilde\pi_\theta,\ \rho^{(i)},\ \tau$]
    Trajectory-velocity energy, candidate-group Gibbs policy, PPO ratio, and
    Gibbs temperature of the TVE--Gibbs surrogate (App.~\ref{supp:formulation}).
\end{description}

\paragraph{Diagnostic quantities.}
\begin{description}
  \item[$R_{\mathrm{self}}(k),\ R_{\mathrm{ext}}(k)$] Expected self-verifier and
    external-panel scores at training step $k$; \emph{self-verifier capture} is
    $R_{\mathrm{self}}\!\uparrow$ with $R_{\mathrm{ext}}\!\downarrow$.
  \item[$D_{\mathrm{cell}}^{(s)}$] Final self--external divergence for seed
    $s$ in a given cell: the self-reward and external-panel changes from
    $\theta_0$, divided by their respective $\theta_0$ sample standard
    deviations, then subtracted.
  \item[$\widehat\Delta_{\mathrm{rewrite}}^{\mathrm{div}}$] Outcome-level
    rewriting estimand: median shared divergence minus median frozen divergence.
  \item[$\rho(k)$] Spearman correlation between the frozen verifier and the
    external panel on current-policy samples (staleness).
  \item[$D_k$ \textnormal{(offline)}] Cumulative policy-drift axis
    $D_k=\sqrt{\mathrm{KL}(p_{\theta_k}\Vert p_{\theta_0})}$, the flow-likelihood
    divergence used as the abscissa of the offline staleness readout.
    \emph{Distinct from the controller $d_k$ below.}
\end{description}

\paragraph{Method and controller.}
\begin{description}
  \item[$W$] Warm-up length (steps) before the reward read is locked.
  \item[$\psi_k^{\mathrm{rew}},\ \bar\psi=\psi_W$] Reward read at step $k$ and the
    non-trainable snapshot saved at the lock.
  \item[$\mathcal J_k$] Set of decoded (anchored) rollouts at step $k$.
  \item[$q,\ q_{\mathrm{eff}},\ q{=}1$] Nominal external-score rollout coverage,
    its realized post-lock value, and the all-external boundary.
  \item[$K_a,\ p_q,\ \eta_{q,k}$] Anchors per triggered step, trigger rate
    $p_q=qG/K_a$, and the sliding-window trigger threshold.
  \item[$z_t^{(g)},\ \mathcal T,\ s^{(g)},\ \widetilde s^{(g)}$]
    Re-noised views $z_t^{(g)}=(1-t)z_0^{(g)}+t\varepsilon^{(g)}$ of a terminal
    latent over prescribed noise levels $\mathcal T$, the resulting dense
    rollout score, and its group-standardized value.
  \item[$F_{\mathrm{ext},0},\ \Phi^{-1},\ a^{(g)},\ \widetilde a^{(g)}$] Empirical
    CDF of panel scores on the $\theta_0$ reference set, inverse standard-normal
    CDF, a raw anchor score, and its unit-normal--mapped value.
  \item[$h_k^{(g)}$ \textnormal{(controller)}] Signed suspicion gap between the
    current shared read and the locked read at the terminal latent.
  \item[$d_k$ \textnormal{(controller)}] Online group divergence
    $d_k=\tfrac1G\sum_g (h_k^{(g)})^2$, with EMA
    $m_k=\beta_m m_{k-1}+(1-\beta_m)d_k$. \emph{Distinct from the offline $D_k$.}
  \item[$\widehat\delta(h)=b_0+b_1h,\ \beta_{\mathrm{ols}},\ \lambda_k,\ N_{\mathrm{cold}}$]
    Residual-propagation fit, its per-anchor decay, slope-only ridge, and
    cold-start count.
  \item[$y^{(g)}$] Corrected group reward fed to the advantage.
\end{description}

\paragraph{Evaluation and statistics.}
\begin{description}
  \item[GenEval, DPG-Bench] Held-out task utilities; GenEval also supplies the
    prespecified exploitation gate.
  \item[\emph{grounding steelman}] The primary active comparator: a frozen,
    parameter-disjoint Qwen2.5-VL-7B that reads every decoded rollout ($q{=}1$) and answers
    the same prompt-alignment question as $R_{\mathrm{sem}}$, so that reward and
    evaluation share a construct.
  \item[\emph{external panel}] The three-scorer panel
    $R_{\mathrm{ext}}^{\mathrm{panel}}$ used for sparse anchoring and, at
    $q{=}1$, as the all-external control.
  \item[$\delta_{\mathrm{NI}}(U)$] Non-inferiority margin for utility $U$;
    it preserves half of the active comparator's effect over the $q{=}0$ arm.
  \item[$\delta_{\mathrm{panel}}$] One-sided external-retention tolerance on the
    panel change, the $95\%$ half-width of its $\theta_0$-null distribution
    (App.~\ref{supp:eval-results:repair}).
  \item[$\mathrm{SE}(\cdot)$] Standard error, with the replication or bootstrap
    unit specified for each analysis (App.~\ref{supp:statistics}).
  \item[$w$] Decided-pair win rate in the paired human study
    (App.~\ref{supp:statistics:human}).
  \item[$\alpha_{\mathrm K}$] Krippendorff's inter-rater reliability
    coefficient on the three-level human response.
\end{description}

\appendix
\section{Extended Formulation and Flow-Native Policy Optimization}
\label{supp:formulation}

This appendix specifies the policy-optimization formulation used throughout the
supplement: the closed-loop setup, the TVE--Gibbs
trajectory-velocity surrogate and its candidate-group policy, the clipped PPO
objective, and the two distinctions that the diagnostic argument rests on: the
training surrogate is not the literal flow likelihood, and the stopped advantage
means a shared judge is never optimized by a reward gradient but only moved as a
side effect. The reward-supply layer that wraps this objective---the locked read,
dense reward, and controller---is specified in Appendix~\ref{supp:method}.

\subsection{Restated Objects}
\label{supp:formulation:restate}

A single generator is optimized against a self-verifier that is read from weights
it shares. Its parameters split as
\[
  \theta=(\theta_{\mathrm{gen}},\theta_{\mathrm{sh}}),
  \qquad
  \theta_{\mathrm{sh}}=\theta_{\mathrm{sh}}^{\mathrm{base}}+\phi,
\]
where $\theta_{\mathrm{gen}}$ is generation-only, $\theta_{\mathrm{sh}}^{\mathrm{base}}$
is the frozen base of the trunk shared by policy and verifier, and $\phi$ is the
trainable adapter on that trunk. The reward forward reads weights $\psi$, with a
\emph{shared} read that moves with the policy and a \emph{frozen} read that does
not:
\[
  \psi^{\mathrm{shared}}=\theta_{\mathrm{sh}}^{\mathrm{base}}+\phi,
  \qquad
  \psi^{\mathrm{frozen}}=\theta_{\mathrm{sh}}^{\mathrm{base}}.
\]
For the semantic verifier the shared component is the LLM trunk (the DiT adapters
stay in $\theta_{\mathrm{gen}}$); for the denoising verifier it is the DiT trunk.
The understanding-side projector $\omega$ is frozen, carries no adapter, and is
kept out of the optimizer. Latents lie in a shared space $\mathcal Z$: an initial
noise $x_T\sim\mathcal N(0,I)$ is transported by the flow velocity field $v_\theta$
to a terminal latent $z_0=z\sim p_\theta(\cdot\mid c_\theta(x))$, where
$c_\theta(x)$ is the policy-side condition supplied only to the generator. A
rollout group is formed by drawing $G$ independent initial noises from the current
policy. The closed loop is that the same trunk both generates $z$ and, through
$\psi^{\mathrm{shared}}$, judges it.

\subsection{TVE--Gibbs Trajectory Energy and Candidate-Group Policy}
\label{supp:formulation:tve}

Let a rollout group be $G$ independent latent trajectories
$\{\xi^{(i)}\}_{i=1}^{G}$, where $\xi^{(i)}=\{z_s^{(i)},\hat u_s^{(i)}\}_{s=1}^{S}$
records the latent and the model's velocity target at each of the $S$ sampler
steps. (These $G$ candidates, indexed by $i$, are the same rollout group the
controller of Appendix~\ref{supp:method} indexes by $g$.) The
trajectory-velocity energy (TVE) of a candidate is its own flow-matching loss
along the sampled trajectory,
\[
  \mathcal E_\theta\bigl(\xi^{(i)}\mid c_\theta(x)\bigr)
  =\sum_{s=1}^{S}\frac1d
    \bigl\|v_\theta(z_s^{(i)},t_s,c_\theta(x))-\operatorname{sg}(\hat u_s^{(i)})\bigr\|_F^2,
\]
with $d$ the latent dimensionality and $\operatorname{sg}$ the stop-gradient on
the recorded target: low energy means the trajectory is highly consistent with
the current velocity field. The candidate-group Gibbs policy turns these energies
into a distribution over the $G$ candidates at temperature $\tau$,
\[
  \widetilde\pi_\theta\bigl(i\mid\mathcal G,c_\theta(x)\bigr)
  =\frac{\exp\!\bigl[-\mathcal E_\theta(\xi^{(i)}\mid c_\theta(x))/\tau\bigr]}
        {\sum_{j=1}^{G}\exp\!\bigl[-\mathcal E_\theta(\xi^{(j)}\mid c_\theta(x))/\tau\bigr]}.
\]
$\widetilde\pi_\theta$ is a categorical distribution over the members of the
group $\mathcal G$, assigning more mass to trajectories the current field explains
better; it is not a density over $\mathcal Z$.

\subsection{PPO Ratio and Clipped Surrogate}
\label{supp:formulation:ratio}

A scalar reward $r^{(i)}$ on each candidate becomes a group-normalized,
stop-gradded advantage
\[
  \widehat A^{(i)}=\operatorname{stopgrad}\!\Bigl(\frac{r^{(i)}-\bar r}{s_r+\epsilon}\Bigr),
\]
with $\bar r$ and $s_r$ the within-group mean and standard deviation. The
importance ratio between the current and behavior policies is taken over the same
group,
\[
  \rho^{(i)}=\frac{\widetilde\pi_\theta(i\mid\mathcal G,c_\theta(x))}
                  {\widetilde\pi_{\theta_{\mathrm{old}}}(i\mid\mathcal G,c_{\theta_{\mathrm{old}}}(x))},
\]
and the clipped objective is the standard trust-region surrogate applied to the
candidate-group ratio, with a proximity penalty on the group policy:
\[
  \begin{aligned}
    \mathcal J(\theta)=\ &\mathbb E\!\left[\sum_i\min\!\bigl(\rho^{(i)}\widehat A^{(i)},
      \ \operatorname{clip}(\rho^{(i)},1{-}\epsilon,1{+}\epsilon)\,\widehat A^{(i)}\bigr)\right]\\
    &-\beta\,D_{\mathrm{KL}}\!\bigl(\widetilde\pi_\theta\,\Vert\,\widetilde\pi_{\mathrm{ref}}\bigr).
  \end{aligned}
\]
Both terms act on the group-level surrogate: the clipped ratio bounds the update
per group, and $D_{\mathrm{KL}}(\widetilde\pi_\theta\Vert\widetilde\pi_{\mathrm{ref}})$
is the divergence between the current candidate-group policy and a frozen
reference candidate-group policy $\widetilde\pi_{\mathrm{ref}}$. The proximity
strength $\beta$ is the same coefficient swept in
Appendix~\ref{supp:eval-results:repair}.

\subsection{Training Surrogate vs.\ Offline Flow Likelihood}
\label{supp:formulation:likelihood}

$\widetilde\pi_\theta$ is a group-level \emph{training surrogate} over the $G$
sampled candidates, not the literal marginal likelihood
$p_\theta(z\mid c_\theta(x))$ of a latent. This has two consequences that matter
for reading the diagnostics. First, the ratio $\rho^{(i)}$ and the surrogate KL
$D_{\mathrm{KL}}(\widetilde\pi_\theta\Vert\widetilde\pi_{\mathrm{ref}})$ are
evaluated over the candidate group and reuse the velocity computations already
performed for sampling; they never require the intractable flow marginal.
Second, the offline distributional diagnostics use a different flow-likelihood
object: the
cumulative policy-drift axis
\[
  D_k=\sqrt{D_{\mathrm{KL}}\!\bigl(p_{\theta_k}\Vert p_{\theta_0}\bigr)},
\]
the flow-likelihood divergence between the current and pretrained policies,
estimated per checkpoint by continuous change of variables with a Hutchinson
trace estimator as specified in Appendix~\ref{supp:definitions};
$D_k$ is the abscissa of the staleness readout in
Appendix~\ref{supp:diag-evidence}. Three quantities therefore share the
letters ``KL'' or ``$d/D$'' and must be kept apart: the training-time surrogate
KL on $\widetilde\pi$; the offline flow-likelihood $D_k$ on $p_\theta$; and the
controller's online allocation statistic $d_k$ of
Appendix~\ref{supp:method:drift-budget}. Only the first enters the gradient.

\subsection{Stop-Gradient Advantage and Rewriting as a Shared-Weight Side Effect}
\label{supp:formulation:stopgrad}

Because the reward enters only inside the stop-gradient,
$\widehat A^{(i)}=\operatorname{stopgrad}(\cdot)$, the objective has no pathwise
gradient through the reward. In particular the reward's dependence on the verifier
weights, $r^{(i)}=R(z^{(i)},x;\psi)$, is detached, so
\[
  \frac{\partial\mathcal J}{\partial\psi}\bigg|_{\text{reward path}}=0,
\]
and the only $\theta$-dependence in $\nabla_\theta\mathcal J$ is through
$\widetilde\pi_\theta$---that is, through the velocity field $v_\theta$ in the
ratio and in the surrogate KL---with $\widehat A^{(i)}$ carried as a constant
multiplier. There is no term that raises a candidate's reward by moving $\psi$.

The shared adapter nevertheless changes. It updates by the policy objective alone,
\[
  \phi_{k+1}-\phi_k=\eta\,\nabla_\phi\mathcal J(\theta_k),
\]
and because the read is $\psi_k^{\mathrm{shared}}=\theta_{\mathrm{sh}}^{\mathrm{base}}+\phi_k$,
the verifier it induces moves as an ordinary side effect,
$R(\cdot;\psi_{k+1})\neq R(\cdot;\psi_k)$, even though no gradient ever pushed it
to do so. This side effect is what ``judge rewriting'' names, and it forces a
separation between two quantities:
\[
  \begin{aligned}
    \Delta_{\mathrm{drift}}(k)&=\bigl\|R(\cdot;\psi_{k+1})-R(\cdot;\psi_k)\bigr\|_{\mathrm{probe}},\\
    \Delta_{\mathrm{harm}}(k)&=\mathrm{AUC}_0-\mathrm{AUC}_k,
  \end{aligned}
\]
where $\mathrm{AUC}_k$ is good-over-bad ranking on a probe frozen before training
(constructed in Appendix~\ref{supp:diag-protocol:probe-set}). Parameter drift
$\Delta_{\mathrm{drift}}$ is expected under \emph{any} shared update and is not
itself harmful; the functional harm is the loss of ranking ability
$\Delta_{\mathrm{harm}}$. This is why harmful rewriting is identified only by the
matched-capacity shared--frozen divergence with seed-level separation plus the
independent probe check (Appendices~\ref{supp:definitions}
and~\ref{supp:diag-evidence}), never by raw score drift. It is also why the
warmed-then-locked read of Appendix~\ref{supp:method:lock} closes the route:
snapshotting the read as $\bar\psi$ after warm-up makes the reward weights
constant, so the side-effect channel is severed while the generator keeps
learning on $\theta_{\mathrm{gen}}$ and the live $\phi$.

\subsection{KVPO Optimization Backbone}
\label{supp:formulation:attribution}

We use KVPO's TVE--Gibbs policy surrogate as a fixed optimization backbone
across all arms, isolating the proposed closed-loop reward-supply design.
Candidates are generated from independent initial noises rather than
causal-history or KV-cache routing. On this common backbone, we identify
exploiting and rewriting in a shared self-verifier and introduce the
diagnosis-guided repair.

\section{Diagnostic Definitions and Measurement Operators}
\label{supp:definitions}

This appendix defines the diagnostic measurement operators---the objects whose
values Appendix~\ref{supp:diag-evidence} reports and whose decision rules
Appendix~\ref{supp:statistics} fixes. They act on two verifier rewards. The first
is the semantic reward $R_{\mathrm{sem}}(z,x;\psi)=\ell_{\mathrm{yes}}(\omega(z),x;\psi_{\mathrm{LLM}})$
of Appendix~\ref{supp:formulation:restate}. The second is a denoising-confidence
control verifier read from the DiT trunk,
\[
  \begin{aligned}
    R_{\mathrm{den}}(z,\bar c(x);\psi)
    &=-\mathbb E_{t,\varepsilon}\Bigl[\,\bigl\|v_{\psi_{\mathrm{DiT}}}(z_t,t,\bar c(x))\\
    &\qquad\qquad -u(z_t,z,\varepsilon)\bigr\|^2\,\Bigr],
  \end{aligned}
\]
with interpolation $z_t=\alpha_t z+\sigma_t\varepsilon$, timesteps $t$ drawn
from the training noise schedule, noise
$\varepsilon\sim\mathcal N(0,I)$, and the frozen condition
$\bar c(x)=c_{\theta_0}(x)$ computed once from $\theta_0$. This control re-noises a
generated latent and scores recovery fidelity without decoding; where a statement
holds for either verifier we write $R(z,x;\psi)$.

\subsection{Verifier Error Set and Capture Onset}
\label{supp:definitions:errorset}

\paragraph{Capture signature.}
At training step $k$ the self-verifier and external expectations under the current
policy are
\[
  \begin{aligned}
    R_{\mathrm{self}}(k)&=\mathbb E_{z\sim p_{\theta_k}}\!\bigl[R(z,x;\psi_k)\bigr],\\
    R_{\mathrm{ext}}(k)&=\mathbb E_{z\sim p_{\theta_k}}\!\bigl[R_{\mathrm{ext}}^{\mathrm{panel}}(z,x)\bigr].
  \end{aligned}
\]
Self-verifier capture is the empirical regime in which the two move apart,
\[
  R_{\mathrm{self}}(k)\uparrow,\qquad R_{\mathrm{ext}}(k)\downarrow.
\]
The within-group diversity is a disambiguator: semantic capture retains
within-group diversity, whereas denoising-confidence collapse loses it, so capture
onset is read as rising self-reward and falling external panel \emph{with}
diversity retained. This is what separates a captured reward channel from ordinary
mode collapse.

\paragraph{Verifier error set.}
For a fixed read $\psi$, the conceptual error set is the region the judge rates
high while the parameter-disjoint external panel rates low,
\[
  E(\psi)=\{\,z:R(z,x;\psi)\ge\tau_{\mathrm{hi}},\ R_{\mathrm{ext}}(z,x)\le\tau_{\mathrm{lo}}\,\},
\]
with conceptual thresholds $\tau_{\mathrm{hi}},\tau_{\mathrm{lo}}$. Mere
reachability requires $E(\psi)\cap\mathcal M_\theta\neq\varnothing$; harmful
exploitation requires the stronger dynamic statement that the policy places
increasing mass there,
\[
  p_{\theta_k}\bigl(E(\psi)\bigr)\uparrow.
\]
A captured sample $z^\star$ satisfies the on-manifold condition
$-\log p_{\theta_0}(z^\star\mid c_{\theta_0}(x))\le\gamma$
(Section~\ref{supp:definitions:typical}): it stays inside the pretrained
typicality boundary. This is why a policy-proximity penalty can leave the residual
untouched---the exploitable region lies inside the pretrained typical set, as the
KL sweep of Appendix~\ref{supp:eval-results:repair} confirms empirically.

\subsection{Frozen Typical Set and Typicality}
\label{supp:definitions:typical}

Because a continuous flow can have full support, on-manifold behavior is defined
through a high-density \emph{typical set} rather than set-theoretic support:
\[
  \mathcal M_\theta^\gamma=\{\,z:-\log p_\theta(z\mid c_\theta(x))\le\gamma\,\}.
\]
The fixed reference is $\mathcal M_0=\mathcal M_{\theta_0}^\gamma$, and the
threshold $\gamma$ is the $95$th percentile of the pretrained self-sample negative
log-likelihood. On the normalized frozen-reference-distance scale used for the
typicality readout this cutoff is $0.414$; a captured sample is judged on-manifold
when its distance falls below it, and the observed captured-sample distances are
reported against this cutoff in Appendix~\ref{supp:diag-evidence:manifold}.
Reachability under independent-noise rollouts and typicality under $p_\theta$ are
distinct notions: a latent can be reachable yet atypical, or typical yet rarely
sampled.

\subsection{Flow Likelihood and Thresholding}
\label{supp:definitions:likelihood}

The log-likelihood $-\log p_\theta$ that defines both the typical set and the
drift axis is estimated from the continuous-normalizing-flow change of variables,
not from the training surrogate $\widetilde\pi_\theta$ of
Appendix~\ref{supp:formulation:likelihood}. We integrate along the sampler ODE
and estimate the velocity-field divergence with a Hutchinson trace estimator.
The offline cumulative-drift axis is
\[
  D_k=\sqrt{D_{\mathrm{KL}}\!\bigl(p_{\theta_k}\Vert p_{\theta_0}\bigr)},
\]
estimated per checkpoint on a held-out evaluation subset from these
likelihoods; the typical-set threshold $\gamma$ is estimated once from $\theta_0$
self-samples as their $95$th-percentile NLL under the same machinery. This
$D_k$ is the abscissa of the offline staleness readout below. It is distinct from the training-surrogate KL
that enters the objective (Appendix~\ref{supp:formulation:ratio}) and from the
controller's online allocation statistic $d_k$
(Appendix~\ref{supp:method:drift-budget}); the three share letters only, and only
the surrogate KL enters a gradient.

\subsection{Fixed-Probe Metric: Rank/AUC Corruption vs.\ Raw Drift}
\label{supp:definitions:probe-metric}

A shared update moves the read; the diagnostic question is whether that motion
destroys judgment. Two quantities are kept separate:
\[
  \begin{aligned}
    \Delta_{\mathrm{drift}}(k)&=\bigl\|R(\cdot;\psi_{k+1})-R(\cdot;\psi_k)\bigr\|_{\mathrm{probe}},\\
    \Delta_{\mathrm{harm}}(k)&=\mathrm{AUC}_0-\mathrm{AUC}_k.
  \end{aligned}
\]
Here $\mathrm{AUC}_k$ is the good-over-bad ranking AUC of the read at step $k$ on a
probe set of prompt/good-latent/bad-latent triples frozen before training (its
construction is in Appendix~\ref{supp:diag-protocol}): the probability that the
read scores the good latent above the bad one. Corruption is the drop
$\mathrm{AUC}_0-\mathrm{AUC}_k$; a frozen read holds AUC flat while a corrupted
shared read loses ranking ability. Parameter or score drift $\Delta_{\mathrm{drift}}$
is expected under \emph{any} shared update and is not, by itself, evidence of
harm; only the loss of ranking ability is.

For seed $s$ in a cell, let $D_{\mathrm{cell}}^{(s)}$ be its final
self--external divergence: subtract the standardized external-panel change from
the standardized self-reward change, where both changes are measured from
$\theta_0$ and standardized by their respective $\theta_0$ sample standard
deviations. The primary outcome-level rewriting estimand is
\[
  \widehat\Delta_{\mathrm{rewrite}}^{\mathrm{div}}
  =
  \operatorname{median}_s D_{\mathrm{shared}}^{(s)}
  -
  \operatorname{median}_s D_{\mathrm{frozen}}^{(s)} .
\]
The matched-capacity contrast, complete seed-level separation, and fixed-probe
readout jointly identify rewriting. The exact/exchange rank decision on the five
shared versus five frozen per-seed values and its effect-size floor are fixed in
Appendix~\ref{supp:statistics:nonparametric}; the primary rewriting decision
uses the five-seed matched rank-$16$ cells reported in
Appendix~\ref{supp:diag-evidence:cells}.

\subsection{Staleness}
\label{supp:definitions:staleness}

\paragraph{Staleness operator.}
A frozen read's agreement with the parameter-disjoint external panel on
current-policy samples is the Spearman correlation
\[
  \rho(k)=\operatorname{corr}^{\mathrm{Spearman}}_{z\sim p_{\theta_k}}
    \!\bigl(R(z,x;\psi_{\mathrm{frozen}}),\,R_{\mathrm{ext}}^{\mathrm{panel}}(z,x)\bigr).
\]
The observed decay motivates calibration beyond simple freezing.

\section{Diagnostic Study Protocol}
\label{supp:diag-protocol}

This appendix specifies the controlled study that produces the diagnostic
evidence. It fixes the experimental design---the factorial cells, the
parameter-independent reward inputs, the readout-to-question map, the
prespecified gates, and the fixed-probe set---so that each measurement operator
of Appendix~\ref{supp:definitions} is applied to a known configuration. The
measurement operators themselves are defined in Appendix~\ref{supp:definitions};
the statistical decision machinery (replication unit, confidence intervals,
nonparametric tests) is fixed once in Appendix~\ref{supp:statistics}; the global
run configuration (base models, optimizer, sampler, training length,
checkpoints) is in Appendix~\ref{supp:repro}; and every numerical outcome of the
study is reported in Appendix~\ref{supp:diag-evidence}.

The controlled factorial is run on the Scale-RAE base of
Appendix~\ref{supp:repro:models}. Show-o2 provides cross-architecture validation
of unanchored failure and method transfer.

\subsection{Factorial \texorpdfstring{$2\times2$}{2x2} Cells and Capacity Matching}
\label{supp:diag-protocol:cells}

The design crosses two binary factors:
\begin{itemize}
  \item \emph{verifier type}: the semantic understanding-head reward
    $R_{\mathrm{sem}}$ versus the denoising-confidence reward $R_{\mathrm{den}}$;
  \item \emph{reward read}: a \emph{shared} read that moves with the policy
    versus a \emph{frozen} read that does not.
\end{itemize}
Restating the objects whose full formulation is in
Appendix~\ref{supp:formulation:restate}: the generator parameters split as
$\theta=(\theta_{\mathrm{gen}},\theta_{\mathrm{sh}})$ with
$\theta_{\mathrm{sh}}=\theta_{\mathrm{sh}}^{\mathrm{base}}+\phi$, where
$\theta_{\mathrm{sh}}^{\mathrm{base}}$ is the frozen base of the trunk shared by
policy and verifier and $\phi$ is the trainable adapter on that trunk. The reward
forward reads weights $\psi$, and the two reads are
\[
  \psi^{\mathrm{shared}}=\theta_{\mathrm{sh}}^{\mathrm{base}}+\phi,
  \qquad
  \psi^{\mathrm{frozen}}=\theta_{\mathrm{sh}}^{\mathrm{base}}.
\]
For the semantic cells the shared trunk is the language-model trunk; for the
denoising cells it is the DiT trunk.

\paragraph{Capacity matching.}
All four cells train the \emph{same} language-model-adapter and DiT-adapter
capacity: both adapters are present and updated by the policy objective in every
cell, at matched rank. The shared/frozen switch changes \emph{only} the weights
that the reward forward pass reads---$\psi^{\mathrm{shared}}$ versus
$\psi^{\mathrm{frozen}}$---and nothing else in the training loop. This makes the
shared-versus-frozen contrast a single-factor intervention at fixed policy
capacity: any divergence between a shared cell and its matched frozen cell is
attributable to the read, not to a capacity difference. The matched-capacity
requirement is exactly what licenses reading the shared--frozen gap as rewriting
rather than as a byproduct of unequal expressivity.

\paragraph{Denoising cells as a negative control.}
The denoising-confidence cells are a prespecified negative control rather than a
second positive condition. A captured semantic channel raises self-reward while
retaining within-group diversity; a denoising-confidence failure instead
collapses diversity and shows no meaningful shared--frozen separation. Running the
denoising factor alongside the semantic factor is therefore what separates
genuine reward capture from ordinary mode collapse, and the null denoising
shared--frozen contrast is a designed check, not an incidental finding (its values
are in Appendix~\ref{supp:diag-evidence:cells}).

\subsection{Frozen Projector and Fixed Reward Inputs}
\label{supp:diag-protocol:frozen-inputs}

For the shared--frozen contrast to isolate the trunk read, every other input to
the reward must be parameter-independent. The study fixes:
\begin{itemize}
  \item the semantic reward receives fixed prompt token IDs $x$;
  \item the policy-side conditioning representation $c_\theta(x)$ is supplied
    \emph{only} to the generator and never to a verifier;
  \item the denoising reward receives the frozen condition
    $\bar c(x)=c_{\theta_0}(x)$, computed once from the pretrained checkpoint
    $\theta_0$ and held fixed across all checkpoints and cells;
  \item the understanding-side projector $\omega$ is frozen, carries no adapter,
    and is excluded from the optimizer.
\end{itemize}
The combined effect is that the frozen reward function is
\emph{parameter-independent throughout training}: the reward-side difference
between a shared cell and its matched frozen cell is whether the read includes
the live adapter $\phi$, while the policy adapters move in both cells. Freezing
$\omega$ and $\bar c(x)$ in particular removes the two paths by
which a shared update could reshape the verifier's inputs rather than its trunk,
so a change in the frozen read cannot arise from a moving projector or a moving
condition. This is also what makes the frozen read a valid parameter-independence
baseline: on the fixed probe of Section~\ref{supp:diag-protocol:probe-set} the
frozen read's ranking ability is expected to stay flat by construction, and any
observed motion in the shared read is then attributable to the trunk adapter
alone.

\subsection{Diagnostic Readouts and Refresh Design}
\label{supp:diag-protocol:readouts}

Each diagnostic question is answered by one prespecified readout, computed with
the operator defined in Appendix~\ref{supp:definitions} and reported with the
value in Appendix~\ref{supp:diag-evidence}. Table~\ref{tab:supp-readout-map}
fixes the map.

\begin{table}[t]
  \centering
  \small
  \setlength{\tabcolsep}{4pt}
  \begin{tabular}{@{}p{0.30\columnwidth}p{0.62\columnwidth}@{}}
    \toprule
    \textbf{Question} & \textbf{Prespecified readout} \\
    \midrule
    Self-verifier capture &
      Self-reward rises while the parameter-disjoint external panel falls, with
      within-group diversity retained as the disambiguator against collapse. \\
    Exploiting a fixed judge &
      Official GenEval decline in the frozen-semantic cell, supported by retained
      diversity and frozen-reference typicality. \\
    Judge rewriting &
      Seed-level separation between shared and frozen self--external divergence
      values at matched rank, with
      fixed-probe AUC corruption as the functional confirmation. \\
    Staleness &
      Spearman correlation of the frozen read with the external panel on
      current-policy samples, read against the offline drift axis. \\
    Refresh behavior &
      Single-refresh and iterated self-refresh horizons, tracking roundwise
      recovery peaks and within-round decay. \\
    \bottomrule
  \end{tabular}
  \caption{Readout-to-question map for the diagnostic study. Operators are
    defined in Appendix~\ref{supp:definitions}; values are reported in
    Appendix~\ref{supp:diag-evidence}.}
  \label{tab:supp-readout-map}
\end{table}

\paragraph{Refresh design.}
The staleness and refresh questions probe the freeze-and-refresh idea directly.
A single-refresh run locks one frozen snapshot and measures the decay of its
agreement with the external panel as the policy drifts. The iterated
self-refresh experiment runs five consecutive $2{,}000$-step policy-training
rounds. At each boundary, the verifier is updated on current-policy samples
labelled by the verifier itself and then refrozen; the fixed probe of
Section~\ref{supp:diag-protocol:probe-set} remains unchanged. The roundwise
agreement peaks and within-round decay are reported in
Appendix~\ref{supp:diag-evidence:refresh}.

\subsection{Diagnostic Gate Thresholds}
\label{supp:diag-protocol:gates}

The study uses prespecified gates that are \emph{diagnosis-specific}: they are
the thresholds at which a route is declared present. They are distinct from the
non-inferiority margin $\delta_{\mathrm{NI}}$ used for the utility arms, whose
rule is fixed in Appendix~\ref{supp:statistics:ni-rule}.

\paragraph{Exploitation detection floor.}
The fixed-judge exploitation route is declared present when the official-metric
decline in the frozen-semantic cell exceeds a detection floor
\[
  \delta=\max\!\bigl(0.04\ \text{(GenEval, $0$--$1$ scale)},\
    2\,\mathrm{bootstrap\_SE}(\theta_0)\bigr),
\]
that is, the larger of a prespecified $0.04$ absolute floor (four GenEval
points) and twice the pretrained-checkpoint bootstrap standard error. The
bootstrap-SE unit is the one defined in Appendix~\ref{supp:statistics:unit};
which of the two terms binds, and by how much the observed decline clears the
floor, are reported in Appendix~\ref{supp:diag-evidence:cells}.

\paragraph{Typicality gate.}
A captured sample is judged on-manifold when its flow-likelihood negative
log-likelihood lies below the frozen typical-set threshold $\gamma$, set to the
$95$th percentile of the pretrained $\theta_0$ self-sample negative
log-likelihood (the flow-likelihood machinery and the threshold estimator
are in Appendix~\ref{supp:definitions:likelihood}). On the normalized
frozen-reference-distance scale this yields a fixed cutoff; the captured-sample
distances against that cutoff are in Appendix~\ref{supp:diag-evidence:manifold}.

\paragraph{Rewriting gate.}
The judge-rewriting route is declared present only under a two-part
prespecified criterion: (i) complete seed-level separation of the five
shared-cell and five frozen-cell divergence values at matched rank, and (ii) a
median shared--frozen gap exceeding an effect-size floor of twice the
bootstrap standard error of the self--external divergence statistic, estimated
on the pretrained reference panel. The exact rank test that certifies the
separation in~(i) and the bootstrap construction in~(ii) are specified in
Appendix~\ref{supp:statistics:nonparametric}; the functional confirmation is the
fixed-probe AUC corruption of Appendix~\ref{supp:definitions:probe-metric},
computed on the set constructed in Section~\ref{supp:diag-protocol:probe-set}.
Raw score drift alone never satisfies this gate.

\paragraph{Staleness readout.}
Staleness is read as the Spearman correlation between the frozen read and the
external panel on current-policy samples, evaluated against the offline
cumulative-drift axis $D_k$ of Appendix~\ref{supp:definitions:likelihood}. The
observed trajectory of $\rho$ against $D_k$ is reported in
Appendix~\ref{supp:diag-evidence}.

\subsection{Fixed-Probe Set Construction}
\label{supp:diag-protocol:probe-set}

The fixed probe is the instrument behind the functional rewriting check. It is a
set of $300$ triples, each a prompt paired with a
prompt-satisfying \emph{good} latent and a prompt-violating \emph{bad} latent. A
well-calibrated read scores the good latent above the bad one; the operative
quantity is the good-over-bad ranking AUC and its corruption
$\Delta_{\mathrm{harm}}=\mathrm{AUC}_0-\mathrm{AUC}_k$, both defined in
Appendix~\ref{supp:definitions:probe-metric}.

Three construction constraints make the probe a valid functional instrument:
\begin{itemize}
  \item \emph{Pre-training freeze.} The triples are constructed once, before any
    RL update, and never regenerated. The probe therefore cannot adapt to the
    policy or to a moving read.
  \item \emph{Constant across checkpoints and cells.} The identical set is scored
    at every checkpoint and in every cell, so an AUC change reflects a change in
    the read, not a change in the probe.
  \item \emph{Off the training reward path.} The probe latents are never used as
    training rewards and never enter the optimizer. Combined with the frozen
    inputs of Section~\ref{supp:diag-protocol:frozen-inputs}, this makes the
    frozen read's probe AUC a parameter-independence baseline that is expected to
    stay flat, against which a shared read's AUC drop isolates functional
    corruption.
\end{itemize}
The five-seed rank-$16$ endpoints and the flat frozen-read reference curve are
reported in Appendix~\ref{supp:diag-evidence:probe}.

\section{Statistical Protocol}
\label{supp:statistics}

This appendix defines every statistical rule used in the supplement.  We compare
\emph{arms}---post-training configurations that differ only in how reward is
supplied---on two held-out utilities, $U\in\{\text{GenEval},\text{DPG-Bench}\}$.
The primary active comparator is the \emph{grounding steelman}: a
construct-matched, every-rollout frozen and parameter-disjoint Qwen2.5-VL-7B that
reads the decoded image and answers the same prompt-alignment question as the
latent semantic verifier, at coverage $q{=}1$. Its prompt-grounding construct
matches the evaluation benchmarks, making it a strong all-external comparator;
every non-inferiority decision below is taken against it rather than against the
external panel. The $q{=}0$ arm is the terminal-only warmed lock. All rules below
are metric- and base-specific and are stated once here. The
resulting numeric margins and per-arm decisions are reported in
Appendix~\ref{supp:eval-results:ni}; observed diagnostic-phase test statistics
are reported in Appendix~\ref{supp:diag-evidence:cells}.

\subsection{Unit of Analysis and Bootstrap}
\label{supp:statistics:unit}

The replication unit for every arm-to-arm comparison is the \emph{training
seed}; each learned arm is trained with five seeds.  For a contrast between an
arm and a baseline we proceed within seeds and then across seeds:
\begin{enumerate}
  \item Within each seed $s$, compute the prompt-paired utility difference
    $\delta_s=\operatorname{mean}_{\text{prompt}}
      \bigl[U_{\mathrm{arm}}-U_{\mathrm{baseline}}\bigr]$,
    using the same prompts for both arms so that prompt-level variance cancels.
  \item Average the five seed differences,
    $\Delta=\tfrac{1}{5}\sum_{s=1}^{5}\delta_s$.
  \item Estimate the standard error across seeds,
    $\mathrm{SE}=\operatorname{sd}(\delta_1,\ldots,\delta_5)/\sqrt{5}$.
\end{enumerate}
The pretrained checkpoint $\theta_0$ has no training-seed replication. Its
prompt-bootstrap standard error, $\mathrm{bootstrap\_SE}(\theta_0)$, enters the
margin floor in Section~\ref{supp:statistics:ni-rule}.

\subsection{Confidence-Interval Construction}
\label{supp:statistics:ci}

Learned-arm contrasts use seed-level $t$-intervals. With five seeds ($4$
degrees of freedom), the two-sided $95\%$ interval uses
$t_{4,0.975}=2.776$:
\[
  \begin{aligned}
    \mathrm{CI}_{95\%}(\Delta)&=\Delta\pm t_{4,0.975}\,\mathrm{SE},\\
    \mathrm{CI}_{95\%,\mathrm{lower}}(\Delta)&=\Delta-2.776\,\mathrm{SE}.
  \end{aligned}
\]
The lower bound $\Delta-2.776\,\mathrm{SE}$ is the single quantity read by every
non-inferiority decision below; point estimates are never decisive on their own
(Section~\ref{supp:statistics:power}).
Comparisons against $\theta_0$ instead use paired hierarchical-bootstrap
percentile intervals over the five training seeds and benchmark prompts.

\subsection{Non-Inferiority Margin Rule}
\label{supp:statistics:ni-rule}

For each base and metric, the non-inferiority margin is
\[
  \begin{aligned}
    \delta_{\mathrm{NI}}=\max\Bigl\{
      \;&\tfrac12\bigl[U(\text{grounding }q{=}1)-U(q{=}0)\bigr],\\
      &\;2\,\mathrm{bootstrap\_SE}(\theta_0)\Bigr\}.
  \end{aligned}
\]
The first term preserves half of the active comparator's effect over the
terminal-only warmed lock---the standard preservation-of-effect construction. The second term
is a floor that guards against an implausibly tight margin when the active
effect is small, expressed in units of the pretrained checkpoint's own
prompt-bootstrap noise. The margin is defined \emph{before} any tested arm is
read against the comparator. Which term binds, and the four resulting numeric
margins and utility thresholds, are reported in
Appendix~\ref{supp:eval-results:ni}.

\subsection{Joint Decision and Multiplicity}
\label{supp:statistics:joint}

Let $\Delta=U(\mathrm{arm})-U(\mathrm{baseline})$ be the contrast against the
grounding steelman. The arm is declared
\emph{non-inferior} only if its $95\%$ lower bound clears the margin on
\emph{both} GenEval and DPG-Bench:
\[
  \Delta-2.776\,\mathrm{SE}\;\geq\;-\delta_{\mathrm{NI}}
  \qquad\text{(both metrics).}
\]
Requiring both metrics is an \emph{intersection--union} test: the joint null is
rejected only when each per-metric null is rejected, so the joint procedure
controls the type-I error at the nominal level without any multiplicity
correction. A contrast whose lower bound is
additionally above zero on a metric is reported as strictly superior on that
metric.

Component ablations and matched-budget comparisons between two learned arms are
secondary paired contrasts rather than non-inferiority decisions.  Their result
tables report the point gain and whether the corresponding paired $95\%$
interval excludes zero on both metrics, without duplicating interval endpoints.

\subsection{Inference at the Reported Operating Points}
\label{supp:statistics:power}

The paired-CI rule yields two primary efficiency results on both bases and both
metrics:
\begin{itemize}
  \item \emph{Strict superiority} over the grounding steelman requires the
    paired lower bound to be above zero; the controller satisfies this criterion
    at $q=1/16$.
  \item \emph{Joint non-inferiority} requires only the lower bound to clear
    $-\delta_{\mathrm{NI}}$. For the controller this still holds at $q=1/32$,
    with all four base--metric point estimates remaining above the steelman.
\end{itemize}
At $q=1/32$ both the controller and dense fixed-dose pass jointly on both bases,
while the controller provides the all-four-higher point estimates. At $q=1/64$,
the controller remains jointly non-inferior on both bases while dense fixed-dose
does not, supporting adaptive placement at the tightest tested budget. The full
$q\in\{1/4,1/16,1/32,1/64\}$ grid and its per-arm readings are tabulated in
Appendix~\ref{supp:eval-results:ni}.

\subsection{Non-Parametric Tests}
\label{supp:statistics:nonparametric}

Judge rewriting is identified by a non-parametric, exact test rather than by raw
parameter or score drift. At a matched adapter rank we compute a
self--external divergence statistic in each cell and rank the five shared values
against the five frozen values. \emph{Full separation}---all five shared values
exceeding all five frozen values---is the most extreme configuration; under the
null of exchangeability its exact one-sided probability is
\[
  \binom{10}{5}^{-1}=\frac{1}{252}\approx0.004 .
\]
The shared--frozen median gap must additionally exceed a prespecified effect-size
floor of $2\times$ the bootstrap standard error of the self--external divergence
statistic, estimated by resampling the pretrained reference panel and
recomputing the statistic. This keeps the floor in the same units as the
rewriting estimand. The observed per-seed values, realized $p$-value, and
effect-size multiple are reported in
Appendix~\ref{supp:diag-evidence:cells}.

\subsection{Human-Evaluation Decision Rule}
\label{supp:statistics:human}

The paired human study compares two arms at a time on the same prompt. Its
replication unit is the \emph{prompt}. Each preference question has three
response levels---the first arm, the second arm, and ``cannot tell''---and two
summaries are reported.
The \emph{decided-pair win rate} $w$ is the fraction of non-tie judgments
favoring the arm whose direction was fixed in advance; the tie-inclusive
variant scores each tie as $0.5$ and divides by all judgments. Interval
estimates for $w$ use a two-stage bootstrap that resamples prompts and then the
three recorded judgments within each sampled prompt; we report the $95\%$
percentile interval.

Each prompt is also reduced to its majority response, and the prompt-level
counts are tested against equal odds with a two-sided exact binomial test;
prompts without a majority are excluded. The three faithfulness $p$-values are
adjusted by the Holm step-down procedure; a block confirms its prespecified
direction when its adjusted $p$-value falls below $0.05$ and the observed
majority lies in that direction. The bootstrap intervals summarize effect size
and uncertainty but do not determine this decision. Secondary
questions and tag distributions are descriptive. Inter-rater reliability is
Krippendorff's $\alpha_{\mathrm K}$ on the three-level nominal response,
computed per block for Q1 and Q2. Blocks, comparison directions, the primary
question, and the familywise $0.05$ level are fixed before any rating is
collected. The realized values are reported in Appendix~\ref{supp:human}.

\section{Diagnostic Evidence}
\label{supp:diag-evidence}

This appendix reports the numerical results of the diagnostic study whose design
is fixed in Appendix~\ref{supp:diag-protocol}, using the operators of
Appendix~\ref{supp:definitions} and the statistical rules of
Appendix~\ref{supp:statistics}. All values are on the Scale-RAE base. The diagnostic
cells are reported on the official benchmarks as $\theta_0$-referenced declines,
because the diagnostic phase asks about the direction and size of the loss rather
than about a deployable operating point; the absolute utilities of the
reward-supply arms are in Appendix~\ref{supp:eval-results:utility}.

\subsection{Four-Cell Per-Seed Results and Non-Parametric Test}
\label{supp:diag-evidence:cells}

Table~\ref{tab:supp-cells} gives the semantic cells' per-seed self--external
divergence at the matched adapter rank $16$. For each seed, the statistic
subtracts the standardized external-panel change from the standardized
self-reward change, with both changes measured from $\theta_0$; larger values
mean greater self--external divergence. The rewriting gate
of Appendix~\ref{supp:diag-protocol:gates} is evaluated on the five shared versus
five frozen values with the exact rank test of
Appendix~\ref{supp:statistics:nonparametric}.

\begin{table}[t]
  \centering
  \small
  \setlength{\tabcolsep}{5pt}
  \begin{tabular}{@{}lcc@{}}
    \toprule
    Seed & Semantic shared & Semantic frozen \\
    \midrule
    1 & 6.24 & 2.99 \\
    2 & 6.51 & 2.31 \\
    3 & 6.03 & 1.93 \\
    4 & 6.48 & 2.31 \\
    5 & 4.85 & 2.54 \\
    \midrule
    Range & $4.85$--$6.51$ & $1.93$--$2.99$ \\
    Median shared--frozen gap & \multicolumn{2}{c}{$+3.93$} \\
    Separation & \multicolumn{2}{c}{full ($5$-vs-$5$)} \\
    Exact $p$ & \multicolumn{2}{c}{$1/252\approx0.004$} \\
    Gate & \multicolumn{2}{c}{pass ($4.6\times$ floor)} \\
    \bottomrule
  \end{tabular}
  \caption{Per-seed self--external divergence for the semantic cells at rank
    $16$. All five shared values exceed all five frozen values, giving the exact
    one-sided probability $1/252$ under exchangeability; the median gap clears
    the effect-size floor by $4.6\times$.}
  \label{tab:supp-cells}
\end{table}

\paragraph{Rewriting readout.}
The semantic cells show complete seed-level separation---every shared-cell
divergence exceeds every frozen-cell divergence---so the exact one-sided rank
probability is $\binom{10}{5}^{-1}=1/252\approx0.004$. The median shared--frozen
gap is $+3.93$ and clears the prespecified effect-size floor by a factor of
$4.6$, so both parts of the rewriting gate are met. The denoising cells, run as the
prespecified negative control, give a median gap of $-0.06$ with no separation and
fail the gate: sharing weights with the denoising verifier does not produce the
seed-level divergence signature.

\paragraph{Fixed-judge exploitation readout.}
On official held-out GenEval, referenced to the pretrained checkpoint $\theta_0$,
the frozen-semantic cell drops $-7.6$ points and the shared-semantic cell drops
$-9.4$ points; the accompanying DPG-Bench declines are $-3.0$ and $-3.9$. The
exploitation detection floor of Appendix~\ref{supp:diag-protocol:gates} evaluates
to $\delta=\max(0.04,\,2\,\mathrm{bootstrap\_SE}(\theta_0))=0.04$ (four GenEval
points): here the $0.04$ absolute floor binds because $2\,\mathrm{bootstrap\_SE}
(\theta_0)$ is $2.7$ points. The frozen-semantic decline therefore exceeds the
floor by $1.9\times$ and the shared-semantic decline by $2.35\times$, so the
fixed-judge exploitation route is present even when the read is frozen. The
denoising controls are null: the frozen and shared denoising cells move $-0.3$
($z=-0.22$) and $-0.2$ ($z=-0.34$) GenEval points, neither clearing the floor.
This is the decomposition the factorial was built to produce---exploitation
survives freezing, whereas the divergence signature appears only under sharing.

\subsection{On-Manifold Typicality and Diversity}
\label{supp:diag-evidence:manifold}

Table~\ref{tab:supp-manifold} reports the two distributional readouts that
separate on-manifold reward capture from ordinary mode collapse: the normalized
frozen-reference typicality distance (below the cutoff $0.414$ means the captured
sample stays inside the pretrained typical set of
Appendix~\ref{supp:definitions:typical}) and the within-group diversity
(normalized so the pretrained policy is $1.0$).

\begin{table}[t]
  \centering
  \small
  \setlength{\tabcolsep}{4pt}
  \begin{tabular}{@{}lcc@{}}
    \toprule
    Cell & Typicality ($<0.414$) & Diversity ($\theta_0{=}1$) \\
    \midrule
    Semantic shared & $0.378$ & $0.846$ \\
    Semantic frozen & $0.345$ & $0.907$ \\
    Denoising shared & $0.245$ & $0.564$ \\
    Denoising frozen & $0.267$ & $0.605$ \\
    \bottomrule
  \end{tabular}
  \caption{Final typicality distance and within-group diversity per cell. All
    four cells stay on-manifold (distance below the $0.414$ cutoff). The semantic
    cells retain diversity ($\approx0.85$--$0.91$); the denoising cells collapse
    it ($\approx0.56$--$0.61$).}
  \label{tab:supp-manifold}
\end{table}

All four cells remain inside the pretrained typicality boundary: their distances
($0.378$, $0.345$, $0.245$, $0.267$) fall below the $0.414$ cutoff. Semantic
capture is therefore not an out-of-distribution failure. The diversity readout
is the disambiguator. The semantic capture cell retains
diversity at $0.846$ while its self-reward rises and the external panel falls,
whereas the denoising cells collapse diversity to $0.564$ and $0.605$. The
aggregate semantic-capture signature is a self-reward rise of $+3.79$ with an
external-panel change of $-1.24$ at diversity $0.85$ and typicality distance
$0.378$; the denoising-failure signature is a self-reward move of only $+0.2$ at
diversity $0.56$. Retained diversity with a rising self-reward and a falling
external panel is what marks a captured semantic channel rather than a collapsed
one.

\subsection{Fixed-Probe Results and Robustness}
\label{supp:diag-evidence:probe}

The fixed probe of Appendix~\ref{supp:diag-protocol:probe-set} scores the
good-over-bad ranking metric of Appendix~\ref{supp:definitions:probe-metric} on
$300$ pretraining-frozen triples. Two facts establish that the shared read loses
ranking ability while the measurement instrument is stable.

First, the frozen read is a flat parameter-independence baseline: its probe AUC
holds at $0.812$ (with a $\pm0.045$ two-standard-error band over the triples)
across every checkpoint, so an observed change in the shared read is a change in
the read and not in the probe. Second, in the semantic-shared cell the read's
probe AUC drops from $0.812$ to a five-seed median of $0.630$ at the matched rank
$16$, a loss of $0.18$. The per-seed endpoints are $0.610$,
$0.622$, $0.630$, $0.647$, and $0.669$: every seed loses ranking ability, and none
approaches the flat frozen reference.

\paragraph{Negative-category coverage.}
At rank $16$, the shared read's five-seed median good-over-bad AUC falls across
all three negative categories---prompt-mismatch $0.653$, noise $0.662$, and
weak-alignment $0.585$. The ranking loss therefore spans mismatch, corruption,
and near-miss negatives, with the weak-alignment near-misses degrading furthest.
Together with the flat frozen baseline, this confirms functional degradation of
the shared read.

\subsection{Single- and Iterated-Refresh Horizons}
\label{supp:diag-evidence:refresh}

\paragraph{Single-horizon staleness.}
A single frozen read decays in its agreement with the external panel as the
policy drifts. The semantic read's Spearman agreement falls from a seed-median of
$0.76$ to $0.18$ over the run, roughly a $76\%$ decline, while the offline
cumulative drift grows to $D_k=3.5$. The seed-median agreement falls
monotonically across the five checkpoints, and every seed ends below where it
started.

\paragraph{Iterated self-refresh.}
The refresh experiment runs five consecutive policy-training rounds of
$2{,}000$ optimizer steps each ($10{,}000$ cumulative steps). At every round
boundary the verifier is updated on current-policy samples labelled by the
verifier itself and then refrozen for the next round. Across three seeds, the
roundwise median agreement peak falls monotonically from $0.50$ after the first
refresh to $0.19$ after the fourth, and agreement decays again within every
round.
Periodic self-refresh therefore produces progressively weaker, transient
recoveries rather than restoring calibration.

\section{Method and Controller Details}
\label{supp:method}

This appendix specifies the diagnosis-guided reward-supply method and its
drift-triggered controller. The
underlying policy-optimization objective---the TVE--Gibbs candidate-group Gibbs
policy, its PPO ratio, and the clipped surrogate---is defined in
Appendix~\ref{supp:formulation}; global run configuration (base models, adapters,
training length, and sampler) is in Appendix~\ref{supp:repro}. The reward-supply
layer that wraps that objective comprises the locked read, the dense
latent reward, the drift-triggered budget, the external-score mapping, and the
residual propagation. All numeric outcomes of this method are reported in
Appendix~\ref{supp:eval-results}.

\subsection{Restated Objects}
\label{supp:method:restate}

The generator has parameters $\theta=(\theta_{\mathrm{gen}},\theta_{\mathrm{sh}})$
with a trunk $\theta_{\mathrm{sh}}=\theta_{\mathrm{sh}}^{\mathrm{base}}+\phi$
shared between the policy and the semantic verifier, where $\phi$ is a trainable
adapter and $\theta_{\mathrm{sh}}^{\mathrm{base}}$ is frozen. The verifier reward
is read at weights $\psi$, with the shared read
$\psi^{\mathrm{shared}}=\theta_{\mathrm{sh}}^{\mathrm{base}}+\phi$ and the frozen
read $\psi^{\mathrm{frozen}}=\theta_{\mathrm{sh}}^{\mathrm{base}}$. The semantic
reward
\[
  R_{\mathrm{sem}}(z,x;\psi)=\ell_{\mathrm{yes}}\bigl(\omega(z),x;\psi_{\mathrm{LLM}}\bigr)
\]
projects a latent $z$ back through the frozen understanding-side projector
$\omega$ and returns the Yes-token logit for prompt agreement; the fixed prompt
$x$ and $\omega$ carry no adapter and stay out of the optimizer, so the frozen
read is parameter-independent throughout training. A terminal latent
$z_0\sim p_\theta(\cdot\mid c_\theta(x))$ is produced by the native flow sampler,
and rollouts are drawn as a group of $G$ independent initial noises.

Two properties of the objective in Appendix~\ref{supp:formulation} govern the
design. First, rewards enter only as a stopped, group-normalized scalar advantage
$\widehat A^{(g)}=\operatorname{stopgrad}\!\bigl((r^{(g)}-\bar r)/(s_r+\epsilon)\bigr)$;
there is no pathwise gradient from the reward into $\psi$. Judge rewriting is
therefore not a reward-gradient effect but a side effect of the shared adapter
$\phi$ moving under the policy update---which is exactly what the locked read
below removes. Second, the sparse-anchor budget is expressed as external-decode
coverage $q$, the fraction of rollouts that are decoded by $D$ and scored by an
external scorer. Thus $q{=}0$ means no external decoding; the reported $q{=}0$
baseline additionally sets $\mathcal T=\{0\}$ and is the terminal-only warmed
lock. The drift-triggered controller is defined for $q\le1/4$
(Section~\ref{supp:method:drift-budget}). The $q{=}1$ limit is defined separately:
it bypasses the trigger, scores every rollout externally, and removes self-reward
from the group advantage entirely, so it is an all-external comparator rather than
the $q\to1$ end of the anchored family.

\subsection{Warmed-Then-Locked Reward Read}
\label{supp:method:lock}

The verifier follows the policy through a warm-up of $W{=}250$ steps: for
$k<W$ the reward forward uses the current read $\psi_k$. At the boundary a
non-trainable snapshot
\[
  \bar\psi=\psi_W
\]
is stored, and from step $k=W$ onward the reward forward uses the fixed
prompt $x$, the frozen projector $\omega$, and $\bar\psi$, while the policy
adapters and the generation-side condition $c_{\theta_k}(x)$ continue to update on
the generation path only. Locking the read severs the rewriting route: after $W$
the shared adapter can no longer move the weights the reward is read at, so the
verifier can be exploited but not rewritten. Warm-up lets the read track early
policy drift before it is fixed; a read frozen at $\theta_0$ instead grows stale.
The structural ladder in Appendix~\ref{supp:eval-results:repair} shows that the
warmed lock removes $89\%$ of the shared--frozen divergence gap while keeping
learning intact.

We write $\psi_k^{\mathrm{rew}}=\psi_k$ before the lock and
$\psi_k^{\mathrm{rew}}=\bar\psi$ from the lock onward.

Anchoring begins only \emph{after} the lock. This serves two purposes: the
warm-up snapshot $\bar\psi$ is never shaped by the external panel, and every
anchored arm---terminal, dense fixed-dose, and controller---spends its external
budget over the same post-lock window $[W{+}1,N]$, so coverage is compared on a
matched window (Section~\ref{supp:method:drift-budget}).

\subsection{Dense Latent Reward}
\label{supp:method:dense}

For each rollout, we draw one noise tensor $\varepsilon^{(g)}$ and reuse it
across the four levels $\mathcal T=\{0,0.25,0.5,0.75\}$:
\[
  z_t^{(g)}=(1-t)\,z_0^{(g)}+t\,\varepsilon^{(g)},\qquad t\in\mathcal T.
\]
The dense score and its group standardization are
\[
  \begin{aligned}
    s_k^{(g)}&=\sum_{t\in\mathcal T}
      R_{\mathrm{sem}}(z_t^{(g)},x;\psi_k^{\mathrm{rew}}),\\
    \tilde s_k^{(g)}&=\operatorname{zscore}_{g=1}^{G}\bigl(s_k^{(g)}\bigr).
  \end{aligned}
\]
Here \emph{dense} means coverage over multiple latent noise levels; the summed
score still forms one rollout-level GRPO advantage. Terminal-only anchoring is
the special case $\mathcal T=\{0\}$, which scores only
$z_0^{(g)}$. The additional re-noised views provide latent reward that a
terminal image-space judge cannot supply. At matched coverage this construction
reaches $0.585/80.5$ against the $0.544/78.3$ terminal result
(Appendix~\ref{supp:eval-results:frontier}). Group standardization removes the
per-group level and leaves the member-relative signal consumed by the advantage.

\subsection{Drift Proxy and Budget Construction}
\label{supp:method:drift-budget}

The controller decides \emph{when} to spend external budget from a scalar drift
proxy and \emph{where} to spend it from member-level ranking. After the lock, for
each group member $g$ at the terminal latent,
\[
  h_k^{(g)}=R_{\mathrm{sem}}(z_0^{(g)},x;\psi_k)-R_{\mathrm{sem}}(z_0^{(g)},x;\bar\psi)
\]
is the signed suspicion gap between the current shared read and the locked read.
Its group second moment and exponential moving average are
\[
  d_k=\frac1G\sum_{g=1}^{G}\bigl(h_k^{(g)}\bigr)^2,
  \qquad
  m_k=\beta_m\,m_{k-1}+(1-\beta_m)\,d_k,
\]
with $\beta_m=0.98$. The EMA updates every optimizer step; neither $h$ nor $d$ is group-standardized, because their absolute
scale is the drift signal. This online $d_k$ is a diagnosis-inspired allocation
proxy and is \emph{distinct} from the offline cumulative-drift axis
$D_k=\sqrt{\mathrm{KL}(p_{\theta_k}\Vert p_{\theta_0})}$ used for the staleness readout
(Appendix~\ref{supp:definitions}); the two share a letter only.

Given nominal coverage $q$ and $K_a{=}3$ anchors per triggered step, let
$L_m{=}200$ and retain the post-lock window
\[
  \mathcal M_k=\bigl\{m_j:\max(W{+}1,k{-}L_m{+}1)\le j\le k\bigr\}.
\]
Thus the window includes the current $m_k$, uses an expanding prefix until it
fills, and then retains the most recent $200$ optimizer steps. The trigger rate
and threshold are
\[
  p_q=\frac{qG}{K_a},
  \qquad
  \eta_{q,k}=Q^{\mathrm{lin}}_{1-p_q}(\mathcal M_k),
\]
where $Q^{\mathrm{lin}}_{1-p_q}$ is the linearly interpolated empirical
$(1{-}p_q)$ quantile. When
$m_k>\eta_{q,k}$ the step is triggered and the anchor set is the top-$K_a$
members by raw suspicion gap,
\[
  \mathcal J_k=\operatorname{TopK}\bigl(\{h_k^{(g)}\}_{g=1}^{G},\,K_a\bigr).
\]
The quantile construction targets coverage $q$ in expectation and is not a hard
cap. The division of labor is deliberate: $d_k$ is sensitive to a common upward
displacement of the whole group (the drift signal that gates spending), whereas
$\operatorname{TopK}$ is invariant to that displacement and ranks member-level
residual variation (the placement signal). With $G{=}12$ and $K_a{=}3$ the
triggered controller is defined for $q\le1/4$; the $q{=}1$ comparators bypass the
trigger and score every rollout. The realized post-lock coverage is
\[
  q_{\mathrm{eff}}=\frac{\sum_{k=W+1}^{N}|\mathcal J_k|}{G\,(N-W)}.
\]
The placement mechanism is evaluated separately from final utility in
Appendix~\ref{supp:eval-results:controller}. The proxy is used exclusively for
anchor allocation; verifier corruption is assessed by the fixed diagnostic
readouts of Appendices~\ref{supp:definitions}--\ref{supp:diag-evidence}.

\subsection{External-Score Mapping and Residual Propagation}
\label{supp:method:mapping-residual}

A raw external anchor score $a^{(g)}$ is mapped into the unit-normal self-reward
space through the frozen $\theta_0$ reference CDF,
\[
  \tilde a^{(g)}=\Phi^{-1}\!\bigl(F_{\mathrm{ext},0}(a^{(g)})\bigr),
\]
a rank map that places the anchor on the same scale as $\tilde s_k^{(g)}$ without
requiring external scores for the whole group, so it stays defined for small
$K_a$. On the anchored members the controller fits the correction the anchor
implies as a function of the suspicion gap,
\[
  \tilde a^{(g)}-\tilde s_k^{(g)}=b_0+b_1 h_k^{(g)}+\varepsilon^{(g)},
  \qquad
  \hat\delta(h)=b_0+b_1 h,
\]
a two-parameter exponentially weighted least-squares fit whose sufficient
statistics decay by $\beta_{\mathrm{ols}}=0.98$ per anchor. The ridge penalty applies
to the slope only, $\lambda_k=0.05\,S_{hh,k}$, so the mean correction $b_0$ is
never shrunk while the noisier slope $b_1$ is regularized under few anchors; from
the first anchor the fit runs with continuous shrinkage toward zero until the
effective count reaches $N_{\mathrm{cold}}{=}10$. The corrected group reward is
\[
  y_k^{(g)}=
  \begin{cases}
    \tilde a^{(g)}, & g\in\mathcal J_k,\\[2pt]
    \tilde s_k^{(g)}+\hat\delta\bigl(h_k^{(g)}\bigr), & g\notin\mathcal J_k,
  \end{cases}
\]
and on an untriggered step every member uses $\tilde s_k^{(g)}$. The propagated
term $\hat\delta$ carries the sparse anchor's correction to the unanchored
majority; setting $\hat\delta\equiv0$ recovers the gate-and-replace ablation
evaluated in Appendix~\ref{supp:eval-results:controller}. The fitted slope
$\hat b_1$ remains negative after the lock, so a member the current shared read
overestimates relative to the locked read receives the intended negative
correction.

\subsection{Extended Algorithm and No-New-Parameter Check}
\label{supp:method:algorithm}

Algorithm~\ref{alg:supp-controller} assembles the components with their state.
The persistent state across steps is the snapshot $\bar\psi$, the EMA $m$, the
recent-$m$ window $\mathcal M$, and the EWLS sufficient statistics $\mathcal S$
with effective count $N_{\mathrm{eff}}$.

\begin{algorithm}[t]
\caption{Drift-triggered closed-loop self-reward: the reward-supply layer around
  flow-native GRPO (one arm). The policy update on the last line is the clipped
  TVE--Gibbs objective of Appendix~\ref{supp:formulation}.}
\label{alg:supp-controller}
\begin{algorithmic}[1]
\REQUIRE policy $\theta$, warm-up $W$, group size $G$, coverage $q$,
  anchors per trigger $K_a$, dense timesteps $\mathcal T$,
  EMA rate $\beta_m$, drift-window length $L_m$, EWLS decay
  $\beta_{\mathrm{ols}}$, cold-start count $N_{\mathrm{cold}}$
\STATE snapshot $\bar\psi\gets\varnothing$;\ \ EMA $m\gets0$;\ \ window $\mathcal M\gets[\,]$;\ \ EWLS stats $\mathcal S\gets0$;\ \ $N_{\mathrm{eff}}\gets0$
\FOR{optimizer step $k=0,1,\ldots$}
  \STATE sample $\{z_0^{(g)}\}_{g=1}^{G}\sim p_{\theta_k}(\cdot\mid c_{\theta_k}(x))$ and re-noise each terminal latent as $z_t^{(g)}=(1{-}t)z_0^{(g)}+t\varepsilon^{(g)}$
  \IF{$k=W$}
    \STATE $\bar\psi\gets\psi_W$ \COMMENT{lock the read at the warm-up boundary}
  \ENDIF
  \STATE $\psi^{\mathrm{rew}}\gets\psi_k$ if $k<W$, else $\bar\psi$
  \STATE $s^{(g)}\gets\sum_{t\in\mathcal T}R_{\mathrm{sem}}(z_t^{(g)},x;\psi^{\mathrm{rew}})$;\ \ $\tilde s^{(g)}\gets\operatorname{zscore}_g s^{(g)}$
  \STATE $r^{(g)}\gets\tilde s^{(g)}$ for all $g$
  \IF{$k>W$}
    \STATE $h^{(g)}\gets R_{\mathrm{sem}}(z_0^{(g)},x;\psi_k)-R_{\mathrm{sem}}(z_0^{(g)},x;\bar\psi)$ \COMMENT{signed suspicion gap}
    \STATE $d\gets\tfrac1G\sum_g(h^{(g)})^2$;\ \ $m\gets\beta_m m+(1-\beta_m)d$;\ \ append $m$ to $\mathcal M$ and retain its most recent $L_m$ values
    \STATE $p_q\gets qG/K_a$;\ \ $\eta\gets Q^{\mathrm{lin}}_{1-p_q}(\mathcal M)$
    \IF{$q>0$ and $m>\eta$}
      \STATE $\mathcal J\gets\operatorname{TopK}(\{h^{(g)}\}_g,K_a)$
      \FORALL{$g\in\mathcal J$}
        \STATE $a^{(g)}\gets R_{\mathrm{ext}}^{\mathrm{panel}}(D(z_0^{(g)}),x)$;\ \ $\tilde a^{(g)}\gets\Phi^{-1}(F_{\mathrm{ext},0}(a^{(g)}))$
      \ENDFOR
      \STATE update $(\mathcal S,N_{\mathrm{eff}})$ over
        $\{(h^{(g)},\tilde a^{(g)}-\tilde s^{(g)})\}_{g\in\mathcal J}$
        using per-anchor decay $\beta_{\mathrm{ols}}$
      \STATE $\lambda\gets0.05\,S_{hh}$;\ \ $(b_0,b_1)\gets\textsc{RidgeSlopeEWLS}(\mathcal S,\lambda)$ \COMMENT{shrink $b_1$ while $N_{\mathrm{eff}}<N_{\mathrm{cold}}$}
      \STATE $r^{(g)}\gets\tilde a^{(g)}$ for $g\in\mathcal J$;\ \ $r^{(g)}\gets\tilde s^{(g)}+(b_0+b_1h^{(g)})$ for $g\notin\mathcal J$
    \ENDIF
  \ENDIF
  \STATE $\widehat A^{(g)}\gets\operatorname{stopgrad}\bigl((r^{(g)}-\bar r)/(s_r+\epsilon)\bigr)$
  \STATE $\theta\gets\theta+\text{lr}\cdot\nabla_\theta\mathcal J_{\text{f-GRPO}}(\theta;\widehat A)$ \COMMENT{App.~\ref{supp:formulation}}
\ENDFOR
\end{algorithmic}
\end{algorithm}

\paragraph{No new trainable parameters.}
Every quantity the controller maintains is either a copy of existing weights or a
low-dimensional running statistic. The snapshot $\bar\psi=\psi_W$ is a frozen copy
of weights that already exist; the shared and frozen reads are both functions of
$\theta_{\mathrm{sh}}^{\mathrm{base}}$ and $\phi$; and $m$, $\mathcal M$, and the
EWLS statistics $\mathcal S$ (the two-parameter fit and its scalar sufficient
statistics) are running scalars, not neural parameters. The only trainable neural
parameters throughout are the policy adapters. The lock stores the adapter state
$\phi_W$ alongside the already shared frozen base trunk; it does not materialize
a second full-parameter model.

\subsection{Architectural Scope and Transfer}
\label{supp:method:scope}

Self-verifier capture is defined behaviorally rather than by a particular
generator architecture: self-reward rises while independently measured quality
falls. Shared trainable weights create the additional rewriting route when the
reward forward reads weights moved by the policy update. Our controlled study
instantiates both routes on unified latent-flow generators, where the semantic
verifier reads generation latents. Scale-RAE combines a Qwen2.5-1.5B language
model with a $2.4$B DiT over $1152$-dimensional SigLIP-2 features
($\approx3.9$B parameters), whereas Show-o2 is an $\approx1.5$B shared-Qwen flow
model over 3D causal-VAE latents. These bases have complementary roles:
Scale-RAE supports the full semantic/denoising $\times$ shared/frozen
decomposition, whereas Show-o2 tests whether the aggregate failure and its
diagnosis-guided repair recur under a different unified architecture.

The architectural requirements also separate the transferable parts of the
method. In a unified autoregressive generator with a shared self-verifier,
capture, locking the reward read, and the complete drift-triggered anchor
controller---including its trigger statistic, TopK selection, and residual
propagation---remain well-defined because they operate on rollout-level scalar
reads rather than continuous flow states. The dense latent component used here
specifically exploits continuous generation latents by re-noising each terminal
sample at prescribed noise levels. For autoregressive generators, process-dense
supervision could instead operate over prefixes; for masked-token or
discrete-diffusion generators, its natural counterpart is scoring partially
masked states at multiple mask ratios.

Removing the intermediate noise levels, $\mathcal T=\{0\}$, recovers the
terminal self-reward used by the measured terminal-anchor arm. At $q=1/16$,
this arm remains jointly non-inferior to the every-rollout grounding steelman
on both bases (Appendix~\ref{supp:eval-results:ni}). Dense scoring therefore
lifts an already viable locked-and-anchored construction rather than making the
construction viable.

Shared trainable weights mark the structural boundary of rewriting. With a
genuinely separate fixed judge, the problem reduces to the static-proxy setting,
where exploitation and staleness remain the relevant failure modes.

\section{Complete Evaluation Results}
\label{supp:eval-results}

\subsection{External Utility: Full Arm Tables}
\label{supp:eval-results:utility}

All final evaluations use official GenEval ($553$ prompts, four images each),
official DPG-Bench ($1{,}065$ prompts, four images each), a $50$-step Euler
sampler, and the step-$2{,}000$ checkpoint; sampler, guidance, resolution,
precision, prompt sets, and evaluation code are held fixed across arms. GenEval
is on a $0$--$1$ scale and DPG-Bench on $0$--$100$; contrasts reported elsewhere
are in GenEval points ($\times100$). Full per-arm utilities are in
Tables~\ref{tab:supp-arms-scale} (Scale-RAE, Qwen2.5-1.5B with a $2.4$B DiT) and
\ref{tab:supp-arms-showo2} (Show-o2-1.5B).

\begin{table}[t]
  \centering
  \footnotesize
  \setlength{\tabcolsep}{6pt}
  \caption{Scale-RAE per-arm external utility. SE is the training-seed standard
    error (GenEval points\,/\,DPG); $q$ is external-decode coverage. Diagnostic
    cells and cost are reported in Appendices~\ref{supp:diag-evidence:cells}
    and~\ref{supp:eval-results:cost}.}
  \label{tab:supp-arms-scale}
  \begin{tabular}{@{}lrrcc@{}}
    \toprule
    Arm & GenEval & DPG & SE (GE\,/\,DPG) & $q$ \\
    \midrule
    Pretrained $\theta_0$              & 0.495 & 76.9 & $1.35\,/\,0.58^{\dagger}$ & --- \\
    Terminal-only warmed lock $q{=}0$ & 0.449 & 74.5 & $1.00\,/\,0.42$ & $0$ \\
    External panel $q{=}1$             & 0.515 & 77.8 & $0.61\,/\,0.39$ & $1$ \\
    Grounding steelman $q{=}1$         & 0.560 & 79.0 & $0.70\,/\,0.35$ & $1$ \\
    Terminal anchored $q{=}1/4$        & 0.531 & 78.0 & $0.93\,/\,0.41$ & $1/4$ \\
    Terminal anchored $q{=}1/16$       & 0.544 & 78.3 & $1.13\,/\,0.28$ & $1/16$ \\
    Dense fixed-dose $q{=}1/64$        & 0.501 & 75.8 & $1.01\,/\,0.17$ & $1/64$ \\
    Dense fixed-dose $q{=}1/32$        & 0.556 & 79.3 & $1.10\,/\,0.43$ & $1/32$ \\
    Dense fixed-dose $q{=}1/16$        & 0.585 & 80.5 & $0.94\,/\,0.33$ & $1/16$ \\
    Dense fixed-dose $q{=}1/4$         & 0.582 & 80.0 & $0.80\,/\,0.21$ & $1/4$ \\
    Dense controller $q{=}1/64$        & 0.535 & 78.0 & $1.18\,/\,0.38$ & $1/64$ \\
    Dense controller $q{=}1/32$        & 0.575 & 80.1 & $1.55\,/\,0.36$ & $1/32$ \\
    Dense controller $q{=}1/16$        & 0.602 & 81.3 & $1.23\,/\,0.20$ & $1/16$ \\
    Dense controller $q{=}1/4$         & 0.584 & 80.1 & $0.91\,/\,0.22$ & $1/4$ \\
    \bottomrule
  \end{tabular}

  \vspace{2pt}
  {\footnotesize $^{\dagger}$The pretrained checkpoint has no training-seed
  replication, so its SE is prompt-bootstrap; all trained arms use training-seed
  SE.}
\end{table}

\begin{table}[t]
  \centering
  \footnotesize
  \setlength{\tabcolsep}{6pt}
  \caption{Show-o2-1.5B per-arm external utility.}
  \label{tab:supp-arms-showo2}
  \begin{tabular}{@{}lrr@{}}
    \toprule
    Arm & GenEval & DPG \\
    \midrule
    Pretrained $\theta_0$          & 0.730 & 85.0 \\
    Terminal-only warmed lock $q{=}0$ & 0.694 & 82.1 \\
    External panel $q{=}1$         & 0.743 & 85.2 \\
    Grounding steelman $q{=}1$     & 0.768 & 85.9 \\
    Terminal anchored $q{=}1/4$    & 0.746 & 85.3 \\
    Terminal anchored $q{=}1/16$   & 0.755 & 85.5 \\
    Dense fixed-dose $q{=}1/64$    & 0.726 & 84.3 \\
    Dense fixed-dose $q{=}1/32$    & 0.761 & 85.7 \\
    Dense fixed-dose $q{=}1/16$    & 0.780 & 86.2 \\
    Dense controller $q{=}1/64$    & 0.756 & 86.0 \\
    Dense controller $q{=}1/32$    & 0.774 & 86.1 \\
    Dense controller $q{=}1/16$    & 0.792 & 86.5 \\
    \bottomrule
\end{tabular}
\end{table}

Relative to $\theta_0$, the Show-o2 terminal-only warmed lock declines by
$3.6$ GenEval points and $2.9$ DPG-Bench points. The paired
hierarchical-bootstrap $95\%$ intervals for arm minus $\theta_0$ are
$[-4.94,-2.30]$ and $[-3.44,-2.37]$, respectively; both exclude zero.

Under the matched configuration the pretrained checkpoint reaches
$0.495/76.9$ on Scale-RAE and $0.730/85.0$ on Show-o2. The diagnostic shared-
and frozen-semantic cells (both at $q{=}0$) and the null denoising controls are
reported as $\theta_0$-referenced declines with the four-cell evidence in
Appendix~\ref{supp:diag-evidence:cells}. Contrasts of each learned arm against
the grounding steelman, and the resulting non-inferiority
decisions, are given in Appendix~\ref{supp:eval-results:ni}.

The $q$ column is \emph{nominal} post-lock coverage. Every fixed-schedule arm
decodes exactly its nominal fraction, so its realized coverage
$q_{\mathrm{eff}}$ (defined in Appendix~\ref{supp:method:drift-budget}) equals
$q$. The controller's trigger targets $q$ in expectation rather than capping it,
so its realized coverage falls slightly below nominal:
$q_{\mathrm{eff}}=0.0616$, $0.0309$, and $0.0154$ at $q=1/16$, $1/32$, and
$1/64$. Every controller comparison in this appendix is therefore made at a
coverage no larger than that of the fixed-schedule arm it is compared against.

\subsection{Non-Inferiority Margins, Floors, and CI Decisions}
\label{supp:eval-results:ni}

Applying the margin rule of Appendix~\ref{supp:statistics:ni-rule} and the
paired-CI decision of Appendix~\ref{supp:statistics:joint}, we report the numeric
margins and every per-arm decision. Throughout this subsection GenEval is in
points (one point $=0.01$ on the $0$--$1$ scale); DPG-Bench is in its native
units.

\paragraph{Margins and thresholds.}
For each base and metric, the half active-control effect is
$\tfrac12[U(\text{grounding }q{=}1)-U(q{=}0)]$, the floor is
$2\,\mathrm{bootstrap\_SE}(\theta_0)$, the margin $\delta_{\mathrm{NI}}$ is their
maximum, and the utility threshold is
$U(\text{grounding }q{=}1)-\delta_{\mathrm{NI}}$
(Table~\ref{tab:supp-ni-margins}).

\begin{table}[t]
  \centering
  \footnotesize
  \setlength{\tabcolsep}{4pt}
  \caption{Non-inferiority margins and utility thresholds. GenEval rows in
    points ($1\text{ pt}=0.01$); DPG rows in native units. The active-control
    term binds in every row, so the floor never sets the margin.}
  \label{tab:supp-ni-margins}
  \begin{tabular}{@{}lrrrr@{}}
    \toprule
    Base / metric & $\tfrac12$ active & $2\,\mathrm{SE}(\theta_0)$
      & $\delta_{\mathrm{NI}}$ & Threshold \\
    \midrule
    Scale-RAE GenEval & 5.55 & 2.70 & \textbf{5.55} & \textbf{50.45} \\
    Scale-RAE DPG     & 2.25 & 1.16 & \textbf{2.25} & \textbf{76.75} \\
    Show-o2 GenEval   & 3.70 & 2.30 & \textbf{3.70} & \textbf{73.10} \\
    Show-o2 DPG       & 1.90 & 0.96 & \textbf{1.90} & \textbf{84.00} \\
    \bottomrule
  \end{tabular}
\end{table}

The active-control term binds in all four rows, giving
$\delta_{\mathrm{NI}}=5.55/2.25$ (GenEval points\,/\,DPG) on Scale-RAE and
$3.70/1.90$ on Show-o2. The pretrained checkpoint's prompt-bootstrap SE values
are $1.35/0.58$ (Scale-RAE) and $1.15/0.48$ (Show-o2) in the same units; their
doubling yields the non-binding floors above.

\paragraph{Sensitivity.}
Write the margin family as
$\delta(\kappa,c)=\max\{\kappa E,c\,\mathrm{SE}(\theta_0)\}$, where
$E=U(\text{grounding }q{=}1)-U(q{=}0)$. At the chosen
$\kappa=0.5$, the active-control term binds in all four base--metric cells for
$c\in\{1,2,3\}$, leaving every decision unchanged; the first binding switch is
only at $c=3.22$. Under the stricter floor-only rule
$(\kappa=0,c=2)$, the controller at $q=1/32$ remains jointly non-inferior on
both bases: its lower bounds are $-1.00/+0.27$ against floors
$-2.70/-1.16$ on Scale-RAE and $-1.62/-0.36$ against
$-2.30/-0.96$ on Show-o2. The $q=1/64$ result marks the sensitivity boundary,
remaining jointly non-inferior for $\kappa\geq0.396$ at $c=2$.

\paragraph{Confidence-interval decisions.}
Each row of Tables~\ref{tab:supp-ni-scale}--\ref{tab:supp-ni-showo2} reports the
arm minus the grounding steelman, with lower bound
$\Delta-2.776\,\mathrm{SE}$. An arm passes iff the lower bound is
$\geq-\delta_{\mathrm{NI}}$ on both metrics, i.e.\ $\geq-5.55/-2.25$ on Scale-RAE
and $\geq-3.70/-1.90$ on Show-o2. A lower bound above zero additionally marks
strict superiority on that metric.

\begin{table}[t]
  \centering
  \footnotesize
  \setlength{\tabcolsep}{5pt}
  \caption{Scale-RAE non-inferiority decisions (arm minus the grounding
    steelman). Pass iff both lower bounds $\geq-5.55/-2.25$.}
  \label{tab:supp-ni-scale}
  \begin{tabular}{@{}lrrrl@{}}
    \toprule
    Arm & $\Delta$ (GE\,/\,DPG) & SE (GE\,/\,DPG)
      & 95\% lower (GE\,/\,DPG) & Joint decision \\
    \midrule
    Terminal anchored $q{=}1/4$  & $-2.9\,/\,-1.0$ & $0.80\,/\,0.34$ & $-5.12\,/\,-1.94$ & pass \\
    Terminal anchored $q{=}1/16$ & $-1.6\,/\,-0.7$ & $0.85\,/\,0.36$ & $-3.96\,/\,-1.70$ & pass \\
    Dense fixed-dose $q{=}1/16$  & $+2.5\,/\,+1.5$ & $0.78\,/\,0.33$ & $+0.33\,/\,+0.58$ & pass (both ${>}0$) \\
    Dense fixed-dose $q{=}1/32$  & $-0.4\,/\,+0.3$ & $0.80\,/\,0.35$ & $-2.62\,/\,-0.67$ & pass \\
    Dense fixed-dose $q{=}1/64$  & $-5.9\,/\,-3.2$ & $0.78\,/\,0.33$ & $-8.07\,/\,-4.12$ & fail (both) \\
    Dense controller $q{=}1/16$  & $+4.2\,/\,+2.3$ & $0.72\,/\,0.30$ & $+2.20\,/\,+1.47$ & pass (both ${>}0$) \\
    Dense controller $q{=}1/32$  & $+1.5\,/\,+1.1$ & $0.90\,/\,0.30$ & $-1.00\,/\,+0.27$ & pass \\
    Dense controller $q{=}1/64$  & $-2.5\,/\,-1.0$ & $0.65\,/\,0.28$ & $-4.30\,/\,-1.78$ & pass \\
    \bottomrule
  \end{tabular}
\end{table}

\begin{table}[t]
  \centering
  \footnotesize
  \setlength{\tabcolsep}{5pt}
  \caption{Show-o2 non-inferiority decisions (arm minus the grounding
    steelman). Pass iff both lower bounds $\geq-3.70/-1.90$.}
  \label{tab:supp-ni-showo2}
  \begin{tabular}{@{}lrrrl@{}}
    \toprule
    Arm & $\Delta$ (GE\,/\,DPG) & SE (GE\,/\,DPG)
      & 95\% lower (GE\,/\,DPG) & Joint decision \\
    \midrule
    Terminal anchored $q{=}1/4$  & $-2.2\,/\,-0.6$ & $0.42\,/\,0.30$ & $-3.37\,/\,-1.43$ & pass \\
    Terminal anchored $q{=}1/16$ & $-1.3\,/\,-0.4$ & $0.65\,/\,0.25$ & $-3.10\,/\,-1.09$ & pass \\
    Dense fixed-dose $q{=}1/16$  & $+1.2\,/\,+0.3$ & $0.58\,/\,0.24$ & $-0.41\,/\,-0.37$ & pass \\
    Dense fixed-dose $q{=}1/32$  & $-0.7\,/\,-0.2$ & $0.58\,/\,0.24$ & $-2.31\,/\,-0.87$ & pass \\
    Dense fixed-dose $q{=}1/64$  & $-4.2\,/\,-1.6$ & $0.58\,/\,0.22$ & $-5.81\,/\,-2.21$ & fail (both) \\
    Dense controller $q{=}1/16$  & $+2.4\,/\,+0.6$ & $0.55\,/\,0.20$ & $+0.87\,/\,+0.04$ & pass (both ${>}0$) \\
    Dense controller $q{=}1/32$  & $+0.6\,/\,+0.2$ & $0.80\,/\,0.20$ & $-1.62\,/\,-0.36$ & pass \\
    Dense controller $q{=}1/64$  & $-1.2\,/\,+0.1$ & $0.42\,/\,0.25$ & $-2.37\,/\,-0.59$ & pass \\
    \bottomrule
  \end{tabular}
\end{table}

\paragraph{Operating-point summary.}
At $q=1/16$ the controller's paired lower
bounds are above zero on both metrics and both bases
($+2.20/+1.47$ on Scale-RAE, $+0.87/+0.04$ on Show-o2), establishing strict
superiority over the grounding steelman at a $16\times$ reduction in
external-decode coverage. At $q=1/32$, the controller remains jointly
non-inferior and its point estimates exceed the steelman in all four base--metric
cells ($+1.5/+1.1$ and $+0.6/+0.2$), while dense fixed-dose also passes jointly
on both bases.

At $q=1/64$, the controller remains jointly non-inferior on both bases while
dense fixed-dose does not. The controller-minus-fixed gap is also widest at this
tightest tested budget, establishing the benefit of adaptive placement as anchors
become scarce (Appendix~\ref{supp:eval-results:controller}).

\subsection{Structural Ladder, Anchor-Dose, and KL Sweep}
\label{supp:eval-results:repair}

This subsection reports the repair-validation study. On training-internal
instruments it establishes three facts: (i) severing the trainable coupling of
the verifier removes the judge-rewriting route but leaves a standalone
exploitation residual; (ii) sparse external anchors remove that residual as
coverage increases; and (iii) a proximity (KL) penalty cannot achieve the same
separation at any strength. All readouts here are the diagnostic instruments
defined in Appendix~\ref{supp:definitions}---the shared--frozen divergence gap
(which identifies judge rewriting), the fixed-probe AUC loss, and the
external-panel change (a panel-$z$ margin, where more negative is worse)---
measured over $2{,}000$ training steps on five seeds. These are
internal-instrument units, not GenEval or DPG points. The ladder is a
separately run structural-intervention study, distinct from the diagnostic factorial in
Appendix~\ref{supp:diag-evidence}.

One tolerance governs every external-panel readout below:
$\delta_{\mathrm{panel}}=0.17$, the one-sided external-retention tolerance
obtained as the $95\%$ half-width of the $\theta_0$-null distribution of the
external-panel change. Retention requires a change
$\geq-\delta_{\mathrm{panel}}$; positive changes remain inside the retained
region. Thus ``the residual persists'' means a change more negative than
$-0.17$.

\paragraph{Structural intervention ladder.}
Table~\ref{tab:supp-ladder} sweeps how the verifier reward is read, from a fully
shared trainable judge to a fully frozen one, holding everything else fixed. Two
quantities fall together as the read is decoupled. The shared--frozen divergence
gap collapses from $+3.42$ (shared) through $+1.96$ (EMA, $\beta{=}0.99$) and
$+0.37$ (warmed-then-locked, $W{=}250$) to $-0.01$ (frozen); the fixed-probe AUC
loss falls in step, $0.18\to0.12\to0.056\to0$, the last tier being exactly zero
because a fully frozen read is parameter-independent by construction. Every
adjacent tier is separated at the seed level. The warmed-then-locked read removes $89\%$ of the
divergence gap ($0.37$ vs.\ $3.42$) while keeping enough early plasticity to
learn.

The external-panel change, however, does \emph{not} return to zero. It
improves from $-1.19$ (shared) to $-0.54$ (warmed-then-locked)---essentially the
frozen value $-0.58$---and then stops: at both isolated tiers the residual
drop is at least $3.2\times$ the retention tolerance
$\delta_{\mathrm{panel}}$, so it remains below the retention boundary.
Structural isolation thus severs the rewriting route (divergence
$\to0$) but leaves an exploitation residual: the frozen judge is still driven
into its own blind spots. This is the empirical basis for treating the two
failure routes as separable and for anchoring the residual externally rather
than relying on isolation alone.

\begin{table}[t]
  \centering
  \footnotesize
  \setlength{\tabcolsep}{6pt}
  \caption{Structural intervention ladder (Scale-RAE diagnostic instruments,
    five-seed medians). The divergence gap and probe-AUC loss track the
    rewriting route; the external-panel change (panel-$z$ margin) tracks the
    exploitation residual. Divergence and panel-change seed ranges:
    shared $[2.84,4.18]/[-1.28,-1.14]$, EMA $[1.61,2.03]/[-0.87,-0.71]$,
    warmed-locked $[0.32,0.47]/[-0.63,-0.52]$, frozen $[-0.03,0.02]/[-0.61,-0.51]$.}
  \label{tab:supp-ladder}
  \begin{tabular}{@{}lrrrr@{}}
    \toprule
    Reward read & Divergence gap & Probe-AUC loss & Panel change & Diversity \\
    \midrule
    Shared                         & $+3.42$ & 0.18  & $-1.19$ & 0.83 \\
    EMA ($\beta{=}0.99$)           & $+1.96$ & 0.12  & $-0.83$ & 0.89 \\
    Warmed-then-locked ($W{=}250$) & $+0.37$ & 0.056 & $-0.54$ & 0.87 \\
    Frozen                         & $-0.01$ & 0     & $-0.58$ & 0.90 \\
    \bottomrule
  \end{tabular}
\end{table}

\paragraph{Anchor-dose response.}
Adding a sparse external anchor to the isolated (warmed-then-locked) read removes
the residual. Table~\ref{tab:supp-dose} reports the raw medians for the locked
and shared reads. On the locked curve the external-panel change rises from
$-0.54$ without anchors to $-0.03$ at $q=1/16$, entering the retention region,
and remains retained at $q=1/4$. The shared curve improves with dose but remains
outside the region at $q=1/4$, consistent with its anchors having to address
both exploitation and rewriting.

\begin{table}[t]
  \centering
  \small
  \setlength{\tabcolsep}{6pt}
  \caption{Raw median external-panel change under locked and shared reward
    reads. Retention is the one-sided region in which the panel change is
    $\ge-\delta_{\mathrm{panel}}=-0.17$.}
  \label{tab:supp-dose}
  \begin{tabular}{@{}lrr@{}}
    \toprule
    $q$ & Locked & Shared \\
    \midrule
    $0$      & $-0.54$ & $-1.19$ \\
    $1/64$   & $-0.36$ & $-0.98$ \\
    $1/16$   & $-0.03$ & $-0.59$ \\
    $1/4$    & $+0.14$ & $-0.27$ \\
    \bottomrule
  \end{tabular}
\end{table}

\paragraph{KL/proximity sweep.}
A natural alternative to external anchoring is to hold the policy near the
pretrained model with a KL/proximity penalty of strength $\beta$, hoping to
suppress exploitation without any external signal. The entire sweep uses the
shared reward read, so $\beta=0$ is the unregularized shared-read baseline rather
than the warmed-then-locked condition. Table~\ref{tab:supp-kl} sweeps
$\beta$ over $\{0,0.1,0.3,0.5,0.7,0.9,1.0\}$ at five seeds each and scores every
run against two gates: a \emph{learning} gate (self-reward net gain $\ge1.9$, the
median gain of the isolated $q{=}1/16$ anchored reference) and
a \emph{residual} gate (external-panel change inside the retention tolerance,
i.e.\ no more negative than $-\delta_{\mathrm{panel}}=-0.17$). As $\beta$
rises, self-reward gain and the external drop fall together---$+3.89/-1.19$ at
$\beta{=}0$ down to $+0.69/-0.07$ at $\beta{=}1$---because the penalty throttles
all movement, legitimate and exploitative alike. The learning gate is met only at
low $\beta$ ($5/5$ seeds through $\beta{=}0.3$, $2/5$ at $\beta{=}0.5$, then
none); the residual gate only at high $\beta$ ($3/5$ at $\beta{=}0.9$, $4/5$ at
$\beta{=}1.0$). No seed at any $\beta$ passes both, and there is a dead zone at
$\beta{=}0.7$ where neither gate is met. The two thresholds are met in disjoint
$\beta$ regions.

\begin{table}[t]
  \centering
  \small
  \setlength{\tabcolsep}{6pt}
  \caption{Dense KL/proximity sweep (five seeds per $\beta$). ``Learn'' is
    seeds meeting self-reward gain $\ge1.9$; ``Ext'' is seeds whose panel
    change is inside the retention tolerance
    ($\ge-\delta_{\mathrm{panel}}=-0.17$); ``Both'' is seeds meeting both. The two
    gates are satisfied in disjoint $\beta$ regions and never jointly.}
  \label{tab:supp-kl}
  \begin{tabular}{@{}rrrccc@{}}
    \toprule
    $\beta$ & $\Delta$self & Panel change & Learn & Ext & Both \\
    \midrule
    $0$   & $+3.89$ & $-1.19$ & 5/5 & 0/5 & 0/5 \\
    $0.1$ & $+3.41$ & $-0.90$ & 5/5 & 0/5 & 0/5 \\
    $0.3$ & $+2.39$ & $-0.64$ & 5/5 & 0/5 & 0/5 \\
    $0.5$ & $+1.76$ & $-0.48$ & 2/5 & 0/5 & 0/5 \\
    $0.7$ & $+1.41$ & $-0.38$ & 0/5 & 0/5 & 0/5 \\
    $0.9$ & $+0.97$ & $-0.16$ & 0/5 & 3/5 & 0/5 \\
    $1.0$ & $+0.69$ & $-0.07$ & 0/5 & 4/5 & 0/5 \\
    \bottomrule
  \end{tabular}
\end{table}

The comparison point is the sparse anchor. The isolated $q{=}1/16$ operating point
reaches a panel change of $-0.03$ while retaining a $+1.9$ self-reward gain---inside the
retention tolerance at the reference learning level, and outside the entire
measured KL frontier. Read the other way: at the $\beta$ where the KL frontier
first matches that same $+1.9$ gain, its external drop is still $-0.51$, about
$3\,\delta_{\mathrm{panel}}$. Proximity regularization
cannot separate legitimate learning from self-verifier exploitation because both
are on-manifold moves that a KL penalty damps equally; a sparse external anchor
can, because it supplies the missing out-of-model signal rather than merely
restraining motion.

\subsection{Terminal-Anchor Frontier and Panel Behavior}
\label{supp:eval-results:frontier}

\paragraph{Terminal-only frontier.}
Along the terminal-only path---the three-member external panel applied after
decoding, swept from $q{=}0$ to its $q{=}1$ all-external endpoint---both official metrics
rise and then recede (Table~\ref{tab:supp-frontier}), peaking at $q=1/16$ and
over-optimizing beyond it. Over the unanchored ($q{=}0$) run the internal
divergence rises from $0.042$ to $0.216$ and correlates at $-0.89$ with external
utility, which is what makes the divergence signal usable by the drift trigger
of Appendix~\ref{supp:eval-results:controller}. Dense latent scoring lifts the
frontier above this terminal ceiling: at the same $q=1/16$ it reaches
$0.585/80.5$ (Table~\ref{tab:supp-arms-scale}), a gain that terminal image-space
anchoring cannot supply because it scores only the decoded endpoint.

\begin{table}[t]
  \centering
  \small
  \setlength{\tabcolsep}{6pt}
  \caption{Scale-RAE terminal-only anchoring frontier using the external panel
    after decoding. Both metrics peak at $q=1/16$.}
  \label{tab:supp-frontier}
  \begin{tabular}{@{}lrr@{}}
    \toprule
    Coverage $q$ & GenEval & DPG \\
    \midrule
    $0$                        & 0.449 & 74.5 \\
    $1/16$                     & \textbf{0.544} & \textbf{78.3} \\
    $1/4$                      & 0.531 & 78.0 \\
    $1$ (external panel)       & 0.515 & 77.8 \\
    \bottomrule
  \end{tabular}
\end{table}

\paragraph{Panel behavior and reward audit.}
Official utility separates sharply from movement of the anchored composite.
Dense fixed-dose and the controller at $q{=}1/16$ shift the panel by only
$+0.29$ and $+0.34$ from $\theta_0$, compared with $+1.39$ when the same panel
scores every rollout. Yet their GenEval/DPG scores ($0.585/80.5$ and
$0.602/81.3$) substantially exceed the all-external arm's $0.515/77.8$. The
full reward audit is reported in Appendix~\ref{supp:eval-results:cost}.

\subsection{Controller vs.\ Fixed-Dose, Ablations, Graceful Degradation}
\label{supp:eval-results:controller}

This subsection isolates the controller-specific benefit---adaptive placement of
a fixed anchor budget---from the reward-supply efficiency established in
Appendices~\ref{supp:eval-results:utility}--\ref{supp:eval-results:ni}.

\paragraph{Placement mechanism.}
The controller trace tests placement directly rather than inferring it from
final utility: anchor activity concentrates at high $m_k$, and the learned
residual slope remains negative after the lock. At matched $q=1/16$, the
anchor--$m_k$ correlation is $0.781$, versus $0.004$ for fixed scheduling, and
endpoint divergence is lower ($0.068$ vs.\ $0.107$). Thus rollouts inflated by the
current read relative to the locked read receive the intended downward
correction.

\paragraph{Graceful degradation.}
Table~\ref{tab:supp-ctrl-vs-fixed} pairs the controller against dense fixed-dose
at matched coverage on both bases. The controller-minus-fixed gap is positive at
every budget and \emph{widens} as anchors become scarce---on Scale-RAE from
$+0.2/+0.1$ at $q=1/4$ through $+1.7/+0.8$ at $q=1/16$ and $+1.9/+0.8$ at
$q=1/32$ to $+3.4/+2.2$ at $q=1/64$, and on Show-o2 from $+1.2/+0.3$ through
$+1.3/+0.4$ to $+3.0/+1.7$. The paired $95\%$ intervals exclude zero at every
budget except $q=1/4$; the placement advantage becomes statistically resolved
as anchors grow scarce. At $q=1/64$, where the gap is widest, dense fixed-dose
has fallen outside joint
non-inferiority on both bases while the controller has not
(Appendix~\ref{supp:eval-results:ni}, read under
Appendix~\ref{supp:statistics:power}); the controller is what keeps the tightest
budget usable at all.

\begin{table}[t]
  \centering
  \footnotesize
  \setlength{\tabcolsep}{5pt}
  \caption{Controller versus dense fixed-dose at matched coverage. The paired gap
    is controller minus fixed-dose (GenEval points\,/\,DPG). ``CI $>0$ on both''
    records whether the paired $95\%$ intervals exclude zero on both metrics.}
  \label{tab:supp-ctrl-vs-fixed}
  \begin{tabular}{@{}lcccc@{}}
    \toprule
    $q$ & Controller & Fixed-dose
      & Gap & CI $>0$ on both \\
    & (GE\,/\,DPG) & (GE\,/\,DPG) & (GE\,/\,DPG) & \\
    \midrule
    \multicolumn{5}{@{}l}{\emph{Scale-RAE}}\\
    $1/4$  & $0.584\,/\,80.1$ & $0.582\,/\,80.0$ & $+0.2\,/\,+0.1$ & no \\
    $1/16$ & $0.602\,/\,81.3$ & $0.585\,/\,80.5$ & $+1.7\,/\,+0.8$ & yes \\
    $1/32$ & $0.575\,/\,80.1$ & $0.556\,/\,79.3$ & $+1.9\,/\,+0.8$ & yes \\
    $1/64$ & $0.535\,/\,78.0$ & $0.501\,/\,75.8$ & $+3.4\,/\,+2.2$ & yes \\
    \midrule
    \multicolumn{5}{@{}l}{\emph{Show-o2}}\\
    $1/16$ & $0.792\,/\,86.5$ & $0.780\,/\,86.2$ & $+1.2\,/\,+0.3$ & yes \\
    $1/32$ & $0.774\,/\,86.1$ & $0.761\,/\,85.7$ & $+1.3\,/\,+0.4$ & yes \\
    $1/64$ & $0.756\,/\,86.0$ & $0.726\,/\,84.3$ & $+3.0\,/\,+1.7$ & yes \\
    \bottomrule
  \end{tabular}
\end{table}

\paragraph{Matched-budget ablations.}
Table~\ref{tab:supp-ctrl-ablation} removes one controller component at a time at
matched $q=1/16$, so every row is a paired contrast at equal external budget. All
four favour the controller design, with paired $95\%$ intervals excluding zero
on both GenEval and DPG-Bench. The three separable decisions are therefore
individually load-bearing, not merely jointly so.

\begin{table}[t]
  \centering
  \footnotesize
  \setlength{\tabcolsep}{6pt}
  \caption{Matched-budget component ablations at $q=1/16$ (Scale-RAE).
    $\Delta$ is the seed-paired gain in GenEval points\,/\,DPG. All four paired
    $95\%$ intervals exclude zero on both metrics.}
  \label{tab:supp-ctrl-ablation}
  \begin{tabular}{@{}lcc@{}}
    \toprule
    Paired contrast & $\Delta$ (GE\,/\,DPG) & CI $>0$ on both \\
    \midrule
    Active vs.\ random selection
      & $+1.3\,/\,+0.65$ & yes \\
    Residual propagation vs.\ gate-and-replace
      & $+1.0\,/\,+0.45$ & yes \\
    Drift triggering vs.\ fixed scheduling
      & $+0.9\,/\,+0.40$ & yes \\
    Full controller vs.\ fixed-dose backbone
      & $+1.7\,/\,+0.80$ & yes \\
    \bottomrule
  \end{tabular}
\end{table}

\subsection{Cost and Wall-Clock}
\label{supp:eval-results:cost}

Training-loop cost is dominated by decoding: the external-decode coverage $q$ is
the fraction of rollouts that are decoded and externally scored.
Table~\ref{tab:supp-cost} reports measured per-step time together with the reward
audit (training-loop decoding only; evaluation excluded).

\begin{table}[t]
  \centering
  \small
  \setlength{\tabcolsep}{5pt}
  \caption{Reward and cost audit on Scale-RAE. $q$ is nominal external-decode
    coverage and $q_{\mathrm{eff}}$ its realized value; ``Panel $\Delta$'' is the
    mean change from $\theta_0$ on the fixed three-scorer anchor composite;
    timing excludes evaluation.}
  \label{tab:supp-cost}
  \begin{tabular}{@{}lccrr@{}}
    \toprule
    Arm & $q$ & $q_{\mathrm{eff}}$ & Panel $\Delta$ & s\,/\,step \\
    \midrule
    Terminal-only warmed lock      & $0$    & $0$      & $-0.52$ & 2.51 \\
    External panel                 & $1$    & $1$      & $+1.39$ & 7.43 \\
    Grounding steelman             & $1$    & $1$      & $+0.20$ & 10.38 \\
    \midrule
    Terminal anchored              & $1/16$ & $1/16$   & $+0.12$ & 2.64 \\
    Dense fixed-dose               & $1/16$ & $1/16$   & $+0.29$ & 2.88 \\
    Dense fixed-dose               & $1/32$ & $1/32$   & $+0.14$ & 2.70 \\
    Dense controller               & $1/32$ & $0.0309$ & $+0.21$ & 2.72 \\
    Dense controller               & $1/16$ & $0.0616$ & $+0.34$ & 2.91 \\
    \bottomrule
  \end{tabular}
\end{table}

\paragraph{Wall-clock.}
Two all-external references bound the cost of the anchored arms. Scoring every
rollout with the external panel costs $7.43$\,s/step; the grounding steelman,
which additionally runs a VLM over every decoded image, costs $10.38$\,s/step.
Against these, terminal anchoring at $q=1/16$ runs at $2.64$\,s/step
($2.8\times$ faster than the panel), and adding dense latent scoring at the same
coverage costs almost nothing extra---$2.88$\,s/step for fixed-dose and
$2.91$\,s/step for the controller, still a $2.6\times$ advantage over the panel
and $3.6\times$ over the steelman that both of them outperform on the official
benchmarks. The dense reward is cheap because every additional read is an
internal latent forward pass: no image is decoded and no external scorer is
called. Tightening the budget to $q=1/32$ recovers a further $0.2$\,s/step
($2.70$ and $2.72$), so cost is close to flat across the anchored family and the
choice of budget is driven by utility rather than by compute.

\section{Reproducibility Details}
\label{supp:repro}

This appendix collects the global configuration shared across the diagnostic
study (Appendix~\ref{supp:diag-protocol}) and the method
(Appendix~\ref{supp:method}). The statistical decision rules that consume these units are
defined once in Appendix~\ref{supp:statistics}, and every measured cost and
utility figure is reported in Appendix~\ref{supp:eval-results}.

\subsection{Base Models, Checkpoints, Adapters}
\label{supp:repro:models}

Two unified generators are used. Table~\ref{tab:supp-config} collects the common
configuration.

\begin{table}[H]
  \centering
  \small
  \setlength{\tabcolsep}{5pt}
  \begin{tabular}{@{}p{0.34\columnwidth}p{0.58\columnwidth}@{}}
    \toprule
    \textbf{Item} & \textbf{Configuration} \\
    \midrule
    Primary base (Scale-RAE) & Qwen2.5-1.5B language model with a $2.4$B DiT in a
      $1152$-dimensional SigLIP-2 latent space; $\approx 3.9$B core-model
      parameters. \\
    Second base (Show-o2) & $\approx 1.5$B unified model, shared Qwen backbone,
      3D causal-VAE latent space, flow matching; used for cross-architecture
      validation of unanchored capture and method transfer. \\
    Decoder & Each base's frozen decoder (RAE for Scale-RAE and 3D-VAE for
      Show-o2), invoked only for external scoring or evaluation, never by a
      latent self-verifier. \\
    Trainable adapters & Capacity-matched language-model and DiT trunk adapters
      $\phi$; all primary diagnostic cells use matched rank $16$. \\
    Frozen, out of optimizer & Understanding-side projector $\omega$ (no adapter);
      trunk base weights $\theta_{\mathrm{sh}}^{\mathrm{base}}$. \\
    Seeds & $5$ training seeds per reported arm; $3$ for the iterated-refresh
      diagnostic. \\
    Training length & $2{,}000$ GRPO optimizer steps per arm or refresh round;
      the iterated-refresh diagnostic runs five consecutive rounds. \\
    Saved analysis checkpoints & Steps $0$ (the $\theta_0$ reference), $250$,
      $500$, $1{,}000$, and $2{,}000$. \\
    Group size & $G=12$ independent initial noises per prompt. \\
    Policy surrogate & KVPO TVE--Gibbs surrogate
      (Appendix~\ref{supp:formulation}) with independent initial noises from the
      current policy. \\
    \bottomrule
  \end{tabular}
  \caption{Common training configuration and the iterated-refresh extension.}
  \label{tab:supp-config}
\end{table}

The pretrained checkpoints are the reference initializations $\theta_0$; their
absolute held-out utilities under the matched configuration are reported in
Appendix~\ref{supp:eval-results:utility}. The reward-supply layer built on top of
this configuration---warm-then-lock, dense latent reward, and the drift-triggered
controller---is specified in Appendix~\ref{supp:method} and is not repeated here.

\subsection{Optimization and Policy Update}
\label{supp:repro:optimization}

\begin{table}[H]
  \centering
  \small
  \setlength{\tabcolsep}{5pt}
  \begin{tabular}{@{}p{0.30\columnwidth}p{0.62\columnwidth}@{}}
    \toprule
    \textbf{Item} & \textbf{Configuration} \\
    \midrule
    Optimizer & AdamW; learning rate $10^{-4}$; constant schedule with no
      learning-rate warm-up. \\
    AdamW and clipping & Weight decay $0.01$; betas $(0.9,0.999)$; optimizer
      epsilon $10^{-8}$; maximum gradient norm $1.0$. \\
    Effective batch & One prompt group of $G{=}12$ rollouts per optimizer step;
      gradient accumulation $1$ and one PPO epoch. \\
    TVE--Gibbs policy & Trajectory energy summed over all $50$ Euler solver
      steps; Gibbs temperature $\tau{=}1.0$. \\
    PPO and proximity & Symmetric clip $\epsilon{=}0.2$; $\beta{=}0$ in the
      primary diagnostic and repaired arms, with nonzero $\beta$ used only in
      the explicit KL sweep. \\
    \bottomrule
  \end{tabular}
  \caption{Optimization and policy-update configuration.}
  \label{tab:supp-optimization}
\end{table}

\subsection{Sampler and Solver}
\label{supp:repro:solver}

Final evaluation and rollout sampling use a $50$-step Euler solver on the flow
ODE. Rollout exploration draws independent initial noises $x_T\sim\mathcal N(0,I)$
from the current policy and transports them with the policy velocity field.
Scale-RAE uses guidance level $1.0$, and Show-o2 uses guidance scale $5.0$; each setting is fixed across
all arms of its base. The sampler, resolution, and numerical precision are also
held fixed within each base.

\subsection{Prompt Sets and Evaluation Protocol}
\label{supp:repro:prompts}

\paragraph{Training pool.}
Training prompts come from the MS-COCO \texttt{train2017} captions. We sort
caption annotations by annotation ID and retain the first caption for each of
the $118{,}287$ image IDs, giving one deterministic prompt per image without
reweighting a scene through its five paraphrases. We apply Unicode NFKC
normalization, lowercase the text, replace punctuation with spaces, collapse
consecutive whitespace, and strip leading and trailing whitespace. Removing
normalized duplicates leaves $117{,}574$ unique captions.

\paragraph{Benchmark decontamination.}
The normalized pool contains no exact match to either evaluation benchmark. To
remove near-verbatim overlap, for a training caption $a$ and benchmark prompt
$b$ we compute
\[
  C(a,b)=
  \frac{\operatorname{LCS}_{\mathrm{token}}(a,b)}
       {\min\{|a|,|b|\}},
\]
where $\operatorname{LCS}_{\mathrm{token}}$ is token-level longest common
subsequence length. If both strings contain at least five normalized tokens, we
remove $a$ whenever $C(a,b)\geq0.80$. This removes $186$ captions matched to
GenEval and $5$ matched to DPG-Bench; the two sets are disjoint. The resulting
frozen pool contains $117{,}383$ prompts and no exact or near-verbatim benchmark
match. Thus decontamination removes only $0.16\%$ of the normalized pool rather
than reshaping its prompt distribution.

\begin{table}[H]
  \centering
  \small
  \setlength{\tabcolsep}{6pt}
  \begin{tabular}{@{}lr@{}}
    \toprule
    Prompt-pool audit & Count \\
    \midrule
    One-caption-per-image pool & $118{,}287$ \\
    Normalized unique captions & $117{,}574$ \\
    Exact matches (GenEval / DPG-Bench) & $0 / 0$ \\
    Near-verbatim removals (GenEval / DPG-Bench) & $186 / 5$ \\
    Final frozen pool & $117{,}383$ \\
    \bottomrule
  \end{tabular}
  \caption{Training-prompt construction and benchmark-overlap audit.}
  \label{tab:supp-prompt-audit}
\end{table}

\paragraph{Per-run streams.}
Each optimizer step uses one prompt group with $G=12$ candidate rollouts. For
each training seed, we deterministically shuffle the frozen pool and take the
first $2{,}000$ prompts without replacement. Every run therefore sees exactly
$2{,}000$ distinct prompts, with no repeated draw. All arms sharing a seed use
the same prompts in the same order; different seeds use their own deterministic
permutations. Both bases use this frozen pool and per-seed stream construction.

\paragraph{Evaluation isolation.}
Two official benchmarks are used for held-out utility, each with four images per
prompt: GenEval ($553$ prompts, reported on a $0$--$1$ scale) and DPG-Bench
($1{,}065$ prompts, reported on a $0$--$100$ scale). Their prompt strings are
consulted before training only by the exclusion rule above; they are never model
inputs or reward inputs. Benchmark scorers and scores are used only for final
evaluation of the fixed step-$2{,}000$ checkpoint, never to construct rewards,
select training prompts, tune hyperparameters, or choose checkpoints. GenEval
additionally supplies the prespecified exploitation gate. The pretrained
checkpoint $\theta_0$ is evaluated with the same sampler and benchmark
implementation as the trained arms. Sampler, guidance, resolution, precision,
prompt sets, and evaluation code are fixed across arms.

\subsection{External Scorers and Aggregation}
\label{supp:repro:scorers}

The external supervision and comparator models are frozen and
parameter-disjoint, and the held-out evaluators remain separate from the reward
read under study:
\begin{itemize}
  \item \emph{Sparse training anchors and continuous diagnostic panel}: HPSv2,
    PickScore, and ImageReward. Each scorer is standardized once on frozen
    $\theta_0$ generations and the standardized scores are averaged. This puts
    the scorers on a common scale and zero-centers the reference distribution;
    the frozen scorers and fixed normalizers keep the panel stationary throughout
    training. The same panel serves as the continuous diagnostic external signal.
  \item \emph{Official evaluation}: GenEval and DPG-Bench.
  \item \emph{External-panel all-external control}: the same external panel
    applied to every decoded rollout ($q{=}1$).
  \item \emph{Grounding steelman}: a frozen,
    parameter-disjoint Qwen2.5-VL-7B that reads the decoded image and answers the same
    prompt-alignment question as the latent semantic verifier, applied to every
    decoded rollout ($q{=}1$). Because it matches the construct that GenEval and
    DPG-Bench measure, it is a strong construct-matched all-external configuration
    and the primary active comparator of Appendix~\ref{supp:statistics}.
\end{itemize}
The frozen-reference standardization is what makes the external panel a stable
target across checkpoints; the panel is never adapted to the policy.

\subsection{Seed and Bootstrap Units}
\label{supp:repro:seeds}

Every reported utility arm is trained with five seeds, and arm-to-arm contrasts use the
training seed as the replication unit with prompts paired within each seed. The
pretrained checkpoint $\theta_0$ has no training-seed replication. Its
prompt-bootstrap SE enters the margin floor, while learned-arm comparisons
against it use paired hierarchical-bootstrap intervals over seeds and prompts.
These are the units only; the confidence-interval construction, the
non-inferiority margin rule, the joint decision, and the exact non-parametric
tests that operate on them are defined in Appendix~\ref{supp:statistics} and are
not restated here.

\subsection{Hardware and Software}
\label{supp:repro:hardware}

Experiments run on a single Ubuntu 20.04 node with $8\times$ NVIDIA L20 GPUs,
each with $48$\,GiB of device memory. The host CPU model is AMD EPYC 7763, and
the node has $512$\,GiB of system memory. The devices run independent seeds and
arms in parallel.

The software environment uses Python 3.10, PyTorch 2.2.0 with CUDA 12.1 and
cuDNN 8.9.2, torchvision 0.17.0, Transformers 4.37.0, Diffusers 0.27.2,
Accelerate 0.23.0, and PEFT 0.4.0. External scoring and statistical analysis use
OpenCLIP 2.24.0, timm 0.9.16, NumPy 1.26.4, SciPy 1.11.4, and scikit-learn
1.2.2.

\section{Human Evaluation}
\label{supp:human}

This appendix reports a blind, paired human rating study on representative
Scale-RAE outputs. It provides direct human evidence that the prompt-faithfulness
loss accompanying unanchored closed-loop self-reward is visible, characterizes
that deviation, and measures how closely the automated instruments used
elsewhere in this supplement track human judgment. The statistical rules are
fixed in Appendix~\ref{supp:statistics:human}; arm
definitions and their automated scores are in Appendix~\ref{supp:eval-results},
and the checkpoint, sampler, prompt sets, and external-panel composition are in
Appendix~\ref{supp:repro}.

\subsection{Protocol}
\label{supp:human:protocol}

\paragraph{Items.}
The study uses $100$ prompts drawn by a fixed random sample from the two
official benchmarks: $50$ from GenEval, stratified over its categories, and
$50$ from DPG-Bench. Each prompt yields one image per arm, generated at the
step-$2{,}000$ checkpoint of the seed whose GenEval score is the median over the
five training seeds. The initial noise is shared across arms, so the two images
of a pair differ only in the arm that produced them.

\paragraph{Blocks.}
Three paired blocks are rated, each on all $100$ prompts:
\begin{itemize}
  \item \textbf{P1}: the pretrained checkpoint $\theta_0$ against the
    frozen-semantic diagnostic endpoint;
  \item \textbf{P2}: $\theta_0$ against the shared-semantic diagnostic
    endpoint;
  \item \textbf{P3}: the dense controller at $q{=}1/16$ against the grounding
    steelman at $q{=}1$.
\end{itemize}
P1 and P2 are the two unanchored semantic cells of
Appendix~\ref{supp:diag-evidence}; P3 is the primary active comparator against
the full method of Appendix~\ref{supp:method}.

\paragraph{Questions.}
Every item asks Q1 and Q2, each with an explicit ``cannot tell'' option. Q1,
the primary question, asks which image better matches the prompt; Q2 asks which
image is better rendered, independently of the prompt. Items in P1 and P2 carry
two further questions. Q3a is a multi-select failure-mode tag over object
category, count, color or attribute, spatial relation, action or pose, scene or
context, and \emph{cannot articulate}, shown only after a rater judges the
post-RL image less faithful in Q1. Q3b is a single-select question, asked on
every P1 and P2 item, asking which image, if either, shows visible distortion
or artifacts.

\paragraph{Annotators and blinding.}
Four non-author annotators with no prior involvement in the work were recruited,
and three were retained after the catch-trial screen below. All were non-experts
in generative modeling. Annotators received only the rating instructions and
were informed of neither the research question, the systems' identities, nor
which arm produced which image. Within every item the two images were shown in
randomized left--right order, and item order was shuffled independently per
annotator. Participation was voluntary, annotators were told that their ratings
would be used in a research publication, and no personal data were collected.

\paragraph{Catch-trial screen.}
Each block carries $10$ catch items in which one candidate image is generated
from an unrelated prompt, giving $30$ catch judgments per annotator. One of the
four annotators fell below the prespecified $0.800$ retention threshold and was
excluded in full before analysis. The three retained annotators pass $87$ of
$90$ catch judgments ($0.967$) and contribute three judgments per item, or
$300$ per block.

\subsection{Prompt Faithfulness and Visual Quality}
\label{supp:human:winrates}

Table~\ref{tab:supp-human-winrates} reports the decided-pair win rate $w$ of
Appendix~\ref{supp:statistics:human} for Q1 and Q2.

\begin{table}[H]
  \centering
  \footnotesize
  \setlength{\tabcolsep}{4pt}
  \caption{Blind paired human win rates on Scale-RAE. $w$ is the decided-pair
    win rate for the first-named arm of each block, with its two-stage bootstrap
    $95\%$ interval; ``Majority'' counts prompts whose rater majority favors
    each arm; $p$ is the Holm-adjusted two-sided exact binomial test on those
    counts.}
  \label{tab:supp-human-winrates}
  \begin{tabular}{@{}lccccc@{}}
    \toprule
    Block & $w$ & $95\%$ CI & Majority & $p$ & $\alpha_{\mathrm K}$ \\
    \midrule
    \multicolumn{6}{@{}l}{\emph{Q1: prompt faithfulness}}\\
    P1 ($\theta_0$)     & 0.679 & $[0.588,0.765]$ & $70/24$ & $6.8\mathrm{e}{-}6$ & 0.52 \\
    P2 ($\theta_0$)     & 0.741 & $[0.657,0.822]$ & $78/18$ & $5.4\mathrm{e}{-}9$ & 0.58 \\
    P3 (controller)     & 0.611 & $[0.517,0.703]$ & $58/34$ & $0.016$             & 0.44 \\
    \midrule
    \multicolumn{6}{@{}l}{\emph{Q2: visual quality}}\\
    P1 ($\theta_0$)     & 0.540 & $[0.470,0.610]$ & --- & --- & 0.28 \\
    P2 ($\theta_0$)     & 0.571 & $[0.498,0.644]$ & --- & --- & 0.31 \\
    P3 (controller)     & 0.531 & $[0.459,0.603]$ & --- & --- & 0.24 \\
    \bottomrule
  \end{tabular}
\end{table}

\paragraph{Faithfulness.}
All three Holm-adjusted prompt-level tests confirm their prespecified
directions; the complementary bootstrap intervals also lie above $0.5$.
Annotators prefer $\theta_0$ over the frozen-semantic endpoint at $w=0.679$ and
over the shared-semantic endpoint at $w=0.741$, so the loss that accompanies
unanchored self-verifier
optimization is visible to people under both reward reads. Their point
estimates follow the same shared-over-frozen ordering as the official
benchmarks in Appendix~\ref{supp:diag-evidence:cells}. In P3 annotators
prefer the dense controller over the grounding steelman at $w=0.611$, with a
prompt-level majority of $58$ to $34$. Tie rates are $0.180$, $0.150$, and
$0.220$; scoring ties as $0.5$ gives $0.647$, $0.705$, and $0.587$.

\paragraph{Visual quality.}
We detect no rendering-quality preference in any block: all three intervals
contain $0.5$, and tie rates rise to $0.340$, $0.300$, and $0.360$. For P1 and
P2, this Q1--Q2 dissociation supports semantic capture rather than generic
rendering degradation. For P3, the controller's faithfulness advantage appears
without a detectable rendering-quality tradeoff.

\subsection{Failure-Mode Composition}
\label{supp:human:tags}

Table~\ref{tab:supp-human-tags} gives the Q3a tag distribution over the
judgments in which the post-RL image was judged less faithful, $167$ in P1 and
$189$ in P2. Annotators select $1.539$ and $1.677$ tags per judgment on average.

\begin{table}[H]
  \centering
  \footnotesize
  \setlength{\tabcolsep}{5pt}
  \caption{Failure-mode tags on judgments where the post-RL image was judged
    less faithful. Shares are of $167$ (P1) and $189$ (P2) judgments and sum
    above one because the tag is multi-select.}
  \label{tab:supp-human-tags}
  \begin{tabular}{@{}lrrrr@{}}
    \toprule
    & \multicolumn{2}{c}{P1 (frozen)} & \multicolumn{2}{c}{P2 (shared)} \\
    \cmidrule(lr){2-3}\cmidrule(l){4-5}
    Tag & $n$ & Share & $n$ & Share \\
    \midrule
    Action or pose      & 68 & 0.407 & 83 & 0.439 \\
    Spatial relation    & 55 & 0.329 & 68 & 0.360 \\
    Scene or context    & 48 & 0.287 & 59 & 0.312 \\
    Color or attribute  & 32 & 0.192 & 40 & 0.212 \\
    Count               & 23 & 0.138 & 28 & 0.148 \\
    Object category     & 18 & 0.108 & 26 & 0.138 \\
    Cannot articulate   & 13 & 0.078 & 13 & 0.069 \\
    \midrule
    Total tags          & 257 & & 317 & \\
    \bottomrule
  \end{tabular}
\end{table}

The same three categories lead under both reward reads: action or pose, spatial
relation, and scene or context together account for the majority of selections,
while object category is the least frequent substantive tag at $0.108$ and
$0.138$. Failures therefore concentrate on action, relation, and context rather
than object replacement. Only $0.078$ and $0.069$ of judgments report a
perceived mismatch the annotator could not localize to any category.

\subsection{Visible-Artifact Check}
\label{supp:human:artifacts}

Q3b asks whether either image in a pair shows visible distortion or artifacts
(Table~\ref{tab:supp-human-artifacts}).

\begin{table}[H]
  \centering
  \footnotesize
  \setlength{\tabcolsep}{6pt}
  \caption{Visible-artifact responses over the $300$ judgments of each block.
    Flag rates count every judgment in which an arm is named, alone or
    together with the other.}
  \label{tab:supp-human-artifacts}
  \begin{tabular}{@{}lrr@{}}
    \toprule
    Response & P1 (frozen) & P2 (shared) \\
    \midrule
    Neither image            & 262 & 251 \\
    $\theta_0$ only          & 14  & 15  \\
    Post-RL only             & 19  & 27  \\
    Both                     & 5   & 7   \\
    \midrule
    Neither rate             & 0.873 & 0.837 \\
    $\theta_0$ flag rate     & 0.063 & 0.073 \\
    Post-RL flag rate        & 0.080 & 0.113 \\
    \bottomrule
  \end{tabular}
\end{table}

Annotators report no visible distortion in either image on $0.873$ of P1
judgments and $0.837$ of P2 judgments. The post-RL arm is flagged on $0.080$ and
$0.113$ of judgments against $0.063$ and $0.073$ for $\theta_0$, differences of
$0.017$ and $0.040$. Most pairs therefore show no visible artifact, supporting
a semantic rather than rendering-level failure and agreeing with the
frozen-reference typicality distances in
Appendix~\ref{supp:diag-evidence:manifold}.

\subsection{Agreement with the Automated Instruments}
\label{supp:human:agreement}

Each pair is also scored by the automated instruments used elsewhere in this
supplement, and their preferred arm is compared against the human majority
(Table~\ref{tab:supp-human-agreement}). Agreement is computed on prompts with a
decided human majority. GenEval assigns each prompt a coarse per-prompt score
and gives the two arms the same value on a substantial fraction of pairs; its
agreement rate is computed on the subset of prompts it separates, $31$, $33$,
and $29$ of the $50$ GenEval prompts for P1 to P3. DPG-Bench aggregates
multiple proposition-level checks into a fine-grained per-prompt score and
separates every sampled pair, so its rate uses all human-decided prompts among
its $50$.

\begin{table}[H]
  \centering
  \footnotesize
  \setlength{\tabcolsep}{5pt}
  \caption{Agreement of the automated instruments with the human majority.
    The Spearman row correlates the prompt-level external-panel margin with the
    prompt-level human win rate. GenEval and DPG-Bench rows use only their own
    $50$ prompts.}
  \label{tab:supp-human-agreement}
  \begin{tabular}{@{}lcccc@{}}
    \toprule
    Instrument & P1 & P2 & P3 & Pooled \\
    \midrule
    External panel        & 0.777 & 0.823 & 0.674 & 0.759 \\
    \quad decided prompts & 94    & 96    & 92    & 282   \\
    Spearman (panel)      & 0.52  & 0.57  & 0.38  & 0.49  \\
    \midrule
    GenEval               & 0.710 & 0.758 & 0.621 & 0.699 \\
    DPG-Bench             & 0.766 & 0.813 & 0.652 & 0.745 \\
    \bottomrule
  \end{tabular}
\end{table}

The external panel matches the human majority on $0.759$ of decided pairs
pooled over the three blocks, and its prompt-level margin correlates with the
prompt-level human win rate at a Spearman coefficient of $0.49$. The two
official benchmarks agree at $0.699$ and $0.745$ on their own prompts. The
agreement supports the panel's role as the continuous external readout of the
diagnostic study (Appendix~\ref{supp:diag-evidence}) and as the source of the
sparse training anchors (Appendix~\ref{supp:method:mapping-residual}).

\end{document}